%% file: main.tex
\documentclass[letterpaper,twocolumn,10pt]{article}
\usepackage{usenix}
\usepackage[available,functional]{usenixbadges}

\usepackage{hyperref}
\usepackage{graphicx}
\usepackage{caption}
\usepackage{adjustbox}
\usepackage{mathtools}
\usepackage{amsmath}
\usepackage{paralist}
\usepackage{multirow}
\usepackage{colortbl}
\usepackage{booktabs}
\usepackage[ruled]{algorithm2e}
\usepackage{subcaption}
\usepackage{tcolorbox}
\tcbuselibrary{listings, skins, breakable}
\usepackage{listings}
\usepackage{xcolor}
\usepackage{balance}
\usepackage{url}
\newcommand{\revised}[1]{{\color{black}{#1}}}
\newcommand{\sysname}[0]{{\sc Ava}\xspace}
\newcommand{\eg}{{\it e.g.,}\xspace}
\newcommand{\ie}{{\it i.e.,}\xspace}

\newcommand\blfootnote[1]{%
  \begingroup
  \renewcommand\thefootnote{}\footnote{#1}%
  \addtocounter{footnote}{-1}%
  \endgroup
}

\begin{document}
\title{\sysname: Towards Agentic Video Analytics with Vision Language Models}

\author{\normalfont
Yuxuan Yan$^{1}$,
Shiqi Jiang$^{2,\dagger}$, Ting Cao$^{3}$, Yifan Yang$^{2}$, Qianqian Yang$^{1}$\\\normalfont
Yuanchao Shu$^{1,\dagger}$ , Yuqing Yang$^{2}$, Lili Qiu$^{2}$\\
$^1$\textit{Zhejiang University} \qquad
$^2$\textit{Microsoft Research} \qquad
$^3$\textit{Tsinghua University}
}

\maketitle
\blfootnote{$\dagger$ Corresponding authors.}
\input{tex/abstract}
\input{tex/introduction}

\input{tex/motivation}

\input{tex/overview}

\input{tex/construction}

\input{tex/generation}

\input{tex/implementation}

\input{tex/evaluation}

\input{tex/discussion}

\input{tex/conclusion}

\input{tex/acknowledgement}

\footnotesize
\bibliographystyle{plain}
\bibliography{ref}

\clearpage
\input{tex/appendix}

\end{document}

%% file: tex/abstract.tex
\vspace{-1cm}
\begin{abstract}

AI-driven video analytics has become increasingly important across diverse domains. However, existing systems are often constrained to specific, predefined tasks, limiting their adaptability in open-ended analytical scenarios. The recent emergence of Vision Language Models (VLMs) as transformative technologies offers significant potential for enabling open-ended video understanding, reasoning, and analytics. Nevertheless, their limited context windows present challenges when processing ultra-long video content, which is prevalent in real-world applications. To address this, we introduce \sysname{}, a VLM-powered system designed for open-ended, advanced video analytics. \sysname{} incorporates two key innovations: (1) the near real-time construction of Event Knowledge Graphs (EKGs) for efficient indexing of long or continuous video streams, and (2) an agentic retrieval-generation mechanism that leverages EKGs to handle complex and diverse queries. Comprehensive evaluations on public benchmarks, LVBench and VideoMME-Long, demonstrate that \sysname{} achieves state-of-the-art performance, attaining 62.3\% and 64.1\% accuracy, respectively, significantly surpassing existing VLM and video Retrieval-Augmented Generation (RAG) systems. Furthermore, to  evaluate video analytics in ultra-long and open-world video scenarios, we introduce a new benchmark, \sysname-100. This benchmark comprises 8 videos, each exceeding 10 hours in duration, along with 120 manually annotated, diverse, and complex question-answer pairs. On \sysname-100, \sysname{} achieves top-tier performance with an accuracy of 75.8\%.

The source code of \sysname is available at \url{https://github.com/I-ESC/Project-Ava}. \sysname-100 benchmark could be accessed at \url{https://huggingface.co/datasets/iesc/Ava-100}.

\end{abstract}

%% file: tex/introduction.tex
\section{Introduction}
\label{sec:intro}

\begin{figure}[t]
    \centering
    \includegraphics[width=0.90\linewidth]{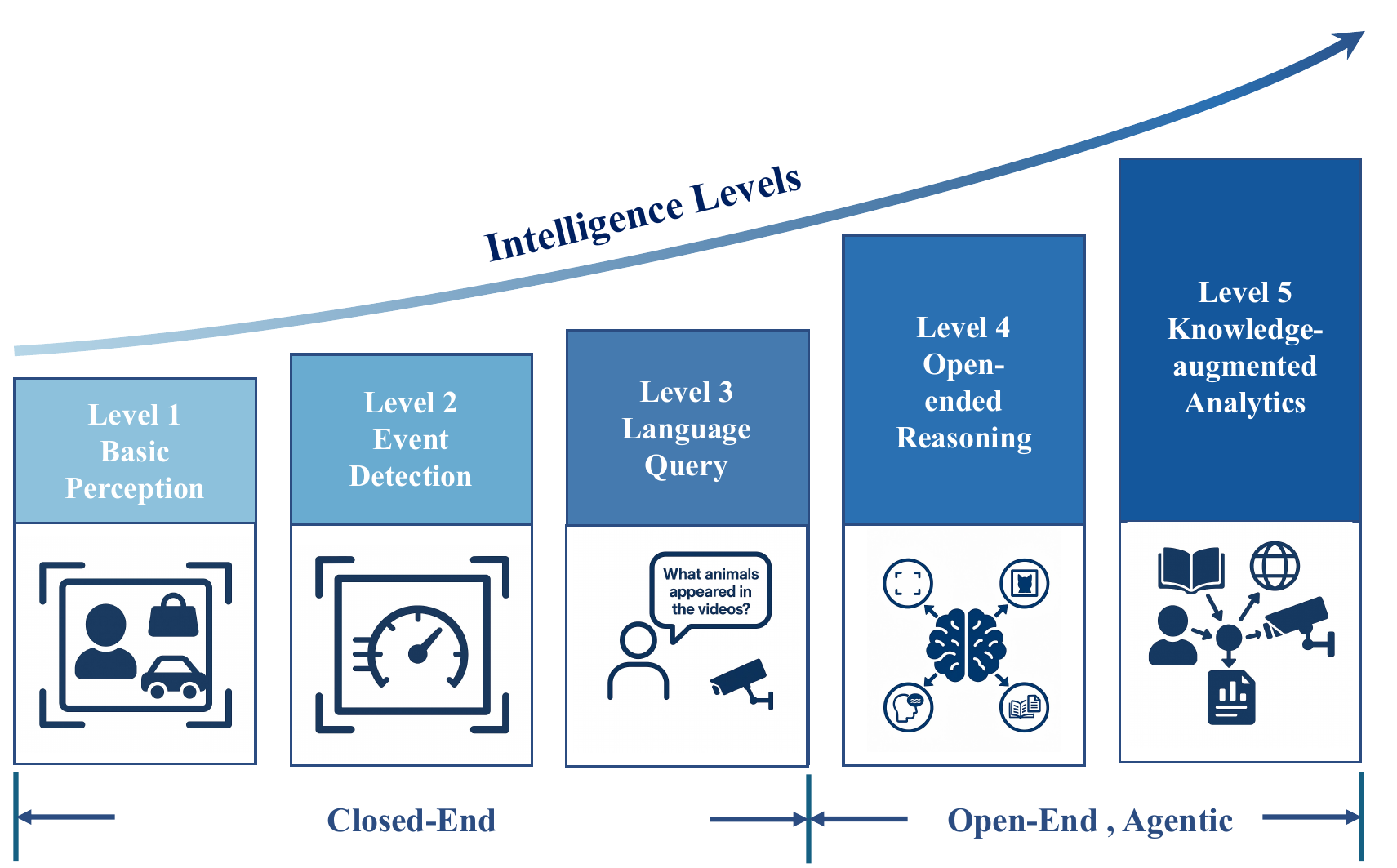}
    \caption{\revised{Intelligence levels of video analytics systems.}}
    \label{fig:l_system}
\end{figure}

Video analytics~\cite{remix_mobicom_21, bhardwaj2022ekya, khani2023recl, 10.1145/3624478} has emerged as a transformative technology across a wide array of domains, such as environment monitoring, intelligent transportation systems, industrial automation, and retail monitoring. \revised{By leveraging the capabilities of deep learning (DL) models, video analytics systems are able to extract patterns, derive meaningful insights, and generate actionable information from video data, thereby facilitating more efficient and precise monitoring, detection, and response to events. The desired features of video analytics systems necessitate a variety of capabilities. As shown in Fig.~\ref{fig:l_system}, We categorize the intelligence levels of both current and prospective video analytics systems into five tiers, designated as L1 through L5:}

\textbf{L1} systems for specific classification, segmentation, and detection using models \eg ResNet~\cite{he2016resnet} and EfficientDet~\cite{tan2020efficientdet} to extract spatial information from video data, including object classes and bounding boxes, etc~\cite{remix_mobicom_21, bhardwaj2022ekya, khani2023recl}.

\textbf{L2} systems go beyond spatial information extraction by enabling causal event detection and analytics, \ie identifying short-term events. They use models like C3D~\cite{tran2015c3d} and ActionFormer~\cite{zhang2022actionformer} to detect and localize events (\eg actions, activities, anomalies) through spatiotemporal modeling.

\revised{\textbf{L3} systems advance beyond L2’s spatiotemporal detection by integrating neural language processing (NLP) capabilities. Leveraging models such as CLIPBERT~\cite{lei2021less}, these systems are able to interpret and respond to natural language queries \eg \textit{"What animals appeared in the videos?"} rather than simply detecting (manually) predefined events. Although query handling remains confined to specific domains, L3 systems significantly enhance user interaction and accessibility in video analytics.}

Despite significant advancements, current video analytics systems~\cite{khani2023recl, bhardwaj2022ekya, zhang2024vulcan, padmanabhan2023gemel, lu2022turbo, sivaraman2024gemino, zhang2021sensei, remix_mobicom_21} primarily focus on L1 to L3 levels. These systems are designed for \emph{closed-end} analytics, relying on specialized models for specific tasks, which limits their flexibility and adaptability. Consequently, we envision L4 and L5 systems to enable \emph{open-end} analytics.

\revised{\textbf{L4} systems represent a significant leap forward by enabling open-ended video comprehension, reasoning, and analytics. These systems support general-purpose video analytics, effectively processing natural language queries and generating contextually appropriate responses, while also facilitating complex, long-term spatiotemporal reasoning. For instance, L4 systems can address inquiries such as: \textit{"What abnormal events occurred in the past ten hours?"}, \textit{"What caused the person to fall?"}, or \textit{"How did the animals behave after appearing on camera?"}}

\revised{\textbf{L5} systems advance beyond L4 by autonomously integrating external public and domain-specific knowledge sources to uncover both explicit and implicit relationships between video content and broader world knowledge. This capability fulfills the ultimate objective of video analytics: deriving profound insights and delivering actionable, automated solutions.}

In this paper, we delve into the development of \textbf{L4 video analytics systems}, leveraging the transformative potential of vision language models (VLMs). By combining vision and language understanding, VLMs enable generalized visual detection and advanced video comprehension, including causal reasoning, key information retrieval, and human-interpretable explanations. The integration of perception, reasoning, and interaction makes VLMs highly adaptable, positioning them as a key foundation for L4 systems.

However, integrating VLMs into video analytics poses significant challenges, primarily due to the limited context window of current VLMs compared to the extensive duration of video sources in typical video analytics scenarios. While L1 to L3 systems handle spatial and short-term causal event detection by processing frames independently or within small sliding windows (\eg a few seconds), L4 systems require collective analysis of related frames for long-term causal detection, summarization, and reasoning. Current VLMs, like QwenVL~\cite{wang2024qwen2}, can process up to 768 frames, covering minutes or hours of video. However, video analytics often involves much longer sources, spanning  hundreds of hours or continuous streams, far beyond the capabilities of existing VLMs.

Recent studies~\cite{shen2024longvu, shen2025longvita, wang2025adaretake} that attempt to extend the context window of VLMs remain inadequate for processing video sources spanning hundreds of hours. Retrieval-augmented generation (RAG) frameworks~\cite{edge2024graphrag, guo2024lightrag, fan2025minirag, lazygraphrag2024} aim to address similar limitations by first retrieving relevant frames from massive contents and then generating final answers. However, these approaches still face significant challenges in handling the video modality (as detailed in \S\ref{sec:evaluation}), leading to both reduced analytics accuracy and substantial computational overhead.

In this paper, we present \sysname, a system that integrates VLMs into video analytics to enable L4 capabilities. The core innovation of \sysname is its efficient indexing mechanism, designed to handle extremely long video sources or unlimited video streams, and by leveraging the index, \sysname effectively retrieves relevant information for a given query and generates accurate, robust responses. Specifically, \sysname introduces the following key features that distinguish it from existing systems: 1) analyzing extremely long videos, spanning hundreds of hours or even unlimited video streams; 2) supporting near-real-time (\eg at more than 1 FPS) index construction and analytics; and 3) handling diverse and complex queries including temporal grounding and reasoning, summarization, event and entity understanding, and key information retrieval.

Particularly, \sysname introduces two key components: \emph{near-real-time index construction} and \emph{agentic retrieval and generation}. During the index construction phase (\S\ref{sec:offline}), we propose event knowledge graphs (EKGs) as an indexing mechanism for video analytics. Unlike traditional knowledge graphs (KGs) used in text-based RAG systems~\cite{edge2024graphrag, guo2024lightrag, lazygraphrag2024}, EKGs represent a flow of insightful events, effectively capturing video dynamics and temporal consistency. \revised{Using a small VLM, such as Qwen2.5-VL-7B}, \sysname extracts information to construct the EKG. To optimize this process, we introduce techniques that enable near-real-time index construction, achieving more than 5 FPS on typical edge servers equipped with 2 $\times$ 4090 GPUs.

In the retrieval and generation phase (\S\ref{sec:online}), \sysname employs an agentic search mechanism instead of directly retrieving information from the constructed index. This approach allows \sysname to proactively retrieve more relevant information by utilizing contextual hints captured within the EKG, enables \sysname to handle complex queries, including summarization and multi-hop reasoning etc. Furthermore, we introduce techniques to enhance \sysname to robustly generate answers based on the retrieved information.

We evaluate \sysname on multiple public video understanding benchmarks, including LVBench~\cite{wang2024lvbench} and VideoMME-Long~\cite{fu2024videomme}. These benchmarks collectively comprise approximately 400 videos and 2,500 questions. We compare \sysname against a diverse range of baselines, \revised{including state-of-the-art (SOTA) VLMs such as GPT-4o~\cite{achiam2023gpt}, Gemini-1.5-pro}~\cite{team2023gemini}, Phi-4-Multimodal~\cite{abdin2024phi3}, Qwen2.5-VL-7B~\cite{wang2024qwen2}, InternVL2.5-8B~\cite{chen2024internvl}, and LLaVA-Video-7B~\cite{zhang2024llavavideo}, as well as typical video-RAG systems such as VideoTree~\cite{wang2024videotree}, VideoAgent~\cite{wang2024videoagent}, DrVideo~\cite{ma2024drvideo}, and VCA~\cite{yang2024vca}. On both benchmarks, \sysname establishes new SOTA performance, achieving 62.3\% on LVBench and 64.1\% on VideoMME-Long, respectively, significantly surpassing baselines by up to 16.9\% and 5.2\%.

In addition to the public video understanding benchmarks, we introduce a new benchmark, \sysname-100, specifically designed to evaluate L4 video analytics tasks. \sysname-100 comprises 8 ultra-long videos, each exceeding 10 hours in duration, and includes a total of 120 manually annotated questions and answers. The videos are carefully curated from typical video analytics scenarios, and the questions cover multiple key categories relevant to video analytics systems. Evaluation results show that \sysname achieves significantly better performance on \sysname-100 compared to various baselines, with improvements of approximately 20.8\%. In summary, we make the following contributions in this paper:

\begin{itemize}
    \item We propose \sysname, the first L4 video analytics system powered by VLMs, to the best of our knowledge.
    \item \sysname introduces near-real-time index construction and agentic retrieval and generation, along with innovative techniques that enable key features for L4 video analytics, including open-ended analytics on extremely long video sources in near-real-time.  
    \item We evaluate \sysname on two widely-used video understanding benchmarks, achieving SOTA performance with 62.3\% on LVBench and 64.1\% on VideoMME-Long.  
    \item Furthermore, we present \sysname-100, a benchmark specifically designed for L4 video analytics systems, where \sysname demonstrates significant improvements, outperforming baselines by approximately 20.8\%.  
\end{itemize}

%% file: tex/motivation.tex
\section{Related Work and Motivation}
\label{sec:motivation}

\subsection{Video Analytics System and VLMs}
\label{subsec:vas_vlm}

The field of video analytics has seen significant advancements in recent years~\cite{khani2023recl, bhardwaj2022ekya, zhang2024vulcan, padmanabhan2023gemel, lu2022turbo, sivaraman2024gemino, zhang2021sensei, jain20sec}. Leveraging emerging deep learning models, modern video analytics systems can extract insightful information, such as object locations or counts, from video streams processed on edge or cloud servers.

Existing video analytics systems predominantly support \emph{closed-end} analytics (L1 to L3 systems as mentioned in \S\ref{sec:intro}), often relying on shallow neural networks to extract predefined, task-specific, and constrained information. For instance, Remix~\cite{remix_mobicom_21} leverages fine-tuned EfficientDet~\cite{tan2020efficientdet} to generate bounding boxes for pedestrians. Consequently, the flexibility and adaptability of video analytics systems are fundamentally limited by the inherent constraints of the specific models they employ.

Recently, VLMs like GPT-4o~\cite{achiam2023gpt}, Gemini~\cite{team2023gemini,team2024gemini}, QwenVL~\cite{wang2024qwen2,bai2025qwen2_5} and Phi~\cite{abdin2024phi3,abdin2024phi4,abouelenin2025phi4mini}, have demonstrated their transformative potential in video analytics tasks. By leveraging the extensive world knowledge embedded in large language models (LLMs), VLMs not only achieve generalized visual grounding but also exhibit advanced video comprehension capabilities, such as zero-shot temporal and spatial reasoning, contextual retrieval, and semantic understanding. More importantly, VLMs enable natural language interaction, allowing users to dynamically query video content. This makes them particularly well-suited for addressing diverse and unstructured \emph{open-ended} analytics in real-world scenarios, towards L4 video analytics systems.

However, adopting VLMs in video analytics systems is far from straightforward. Existing L1 to L3 video analytics systems, which rely on traditional DNNs, typically process each video frame independently. In contrast, VLMs require related frames to be processed collectively to infer causal relationships and temporal dependencies across frames. This shift introduces significant complexity, as existing VLMs are generally capable of handling only minute-level or sub-hour-level videos due to the limited context window inherent in language models.

In real-world video analytics scenarios, the scale of videos to be analyzed is often vastly larger—spanning hundreds of hours or more (\eg monitoring wildlife behavior over an entire month, as illustrated in Fig.~\ref{fig:evg_kg}). This creates a fundamental gap between the capabilities of VLMs and the demands of video analytics systems, as the limited context window length of VLMs directly restricts their ability to process videos of such extensive durations effectively.

\begin{table}[t]
\centering
\resizebox{\columnwidth}{!}{%
\begin{tabular}{cc|cc|cc} 
\toprule[2pt]
\multicolumn{2}{c}{\textbf{Short (1.4 minutes)}} & \multicolumn{2}{c}{\textbf{Medium (9.7 minutes)}} & \multicolumn{2}{c}{\textbf{Long (39.7 minutes)}}  \\ 
\cmidrule{1-6}
Total & Needed & Total & Needed   & Total & Needed \\ 
2144.8 & 12.1 (0.5\%)      & 13924.1     & 68.1 (0.4\%)                   & 66847.1     & 82.3 (0.1\%)                \\
\bottomrule[2pt]
\end{tabular}%
}
\caption{\revised{Only a small portion of the frames are necessary to answer each particular question across the short, medium, and long video subsets of VideoMME~\cite{fu2024videomme} benchmark using Qwen2-VL.}}
\label{tab:few_needed_frames}
\end{table}

\subsection{Long Video Understandings}

Recent research also increasingly focuses on enabling long video understandings~\cite{shen2024longvu, wang2025adaretake}. \revised{Current autoregressive language models inherently have the constrained context window length, therefore efforts have been directed toward reducing the redundancy in video inputs to facilitate the processing of extended video durations.  For instance, LongVU~\cite{shen2024longvu} and AdaRETAKE~\cite{wang2025adaretake} introduce dynamic compression mechanisms that prioritize video content based on its relevance, selectively retaining frames or regions most pertinent to downstream language tasks. Similarly, NVILA~\cite{liu2024nvila} addresses the efficiency-accuracy trade-off by optimizing sampling strategies and resolution to fit within limited token budgets.}

While these approaches have succeeded in increasing the number of frames that models can process and mitigating the constraints of context windows to some extent, they fall short of achieving a fundamental breakthrough. \revised{Given that SOTA LLMs typically support context lengths ranging from 128K to 1M tokens, most existing approaches are thereby restricted to processing video segments of at most one hour in duration~\cite{qian2024streaming}, which is inadequate for the requirements of video analytics. Furthermore, as video length increases, the associated inference cost rises proportionally, thereby compounding the scalability challenges inherent to these systems.}

\subsection{Retrieval Augmented Generation}

\begin{figure*}[t]
   \centering
   \includegraphics[width=.85\linewidth]{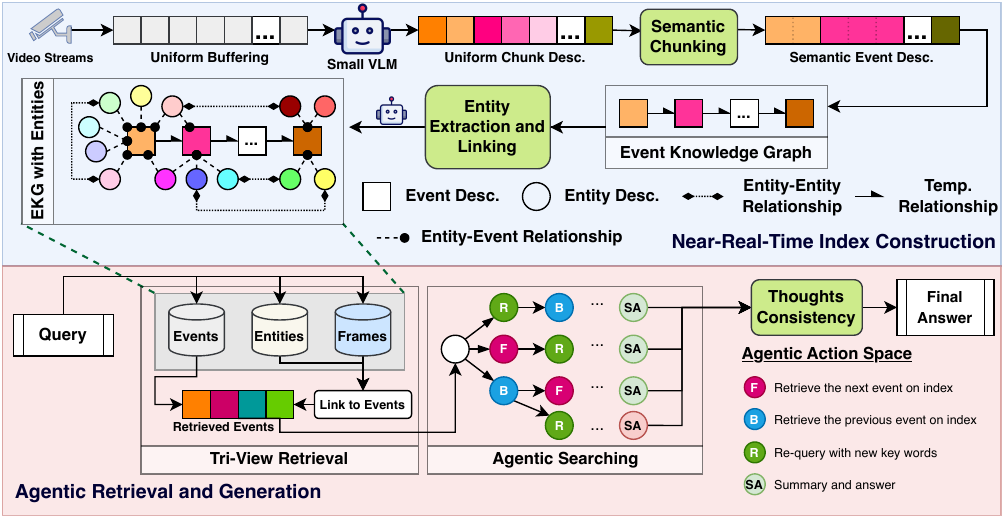}
   \caption{The system overview of \sysname.}
   \label{fig:overview}
\end{figure*}

While the videos to be analyzed may span extensive durations, the frames necessary to respond to a specific query are often limited. To validate this observation, we conducted an experiment on VideoMME~\cite{ma2024drvideo} using Qwen2-VL~\cite{wang2024qwen2}. Specifically, we first identified all questions for which Qwen2-VL produced correct answers by uniformly sampling frames from the videos at a rate of 1 FPS. For these questions, we then determined the minimal set of frames required for the VLM to generate the correct answer by iteratively reducing the number of input frames using a binary search strategy.~\footnote{For example, we initially uniformly sample 100 frames from the video. If the VLM can generate the correct answer based on these frames, we then attempt to reduce the frame set to 50 frames. If 50 frames are still sufficient to produce the correct answer, we further reduce the set to 25 frames. Conversely, if 50 frames are insufficient, we increase the set to 75 frames, iteratively refining the frame selection using binary search strategy.} \revised{The results indicate that, owing to significant temporal redundancy in video data, the frames required to answer a specific query represent only a small fraction of the total frames, as shown in Table~\ref{tab:few_needed_frames}.}

Based on this observation, an intuitive approach would be to first retrieve the relevant frames corresponding to a specific query and then generate the final answer based on these frames, a method commonly referred to as Retrieval-Augmented Generation (RAG). To retrieve potentially relevant frames, a straightforward strategy involves \emph{vectorized retrieval}, where each frame of the video is embedded using a vision-language model, such as CLIP~\cite{radford2021CLIP}. At query time, the embedding of the query is used to retrieve relevant frames by comparing the similarity between the query embedding and the vectorized frames. 

However, the vectorized retrieval method's limitations stem from the detailed information contained in the query. Notably, such an approach struggles to handle query-focused summaries~\cite{edge2024graphrag} (\eg "What happened in the last few hours?") or multi-hop queries~\cite{zhao2024retrieval} (\eg "What did the man do after he opened the fridge?"), as the retrieved frames often fail to capture key contexts that are not explicitly mentioned in the query descriptions.

To enable effective retrieval, recent research has explored two prominent approaches: \emph{video structuring} and \emph{iterative retrieval}. For example, Video-RAG~\cite{luo2024video-rag} structures videos by utilizing various tools to extract information such as automatic speech recognition (ASR), optical character recognition (OCR) results, and object detection outputs. It then applies RAG techniques to the structured information. However, this method is inherently constrained by the tools employed for video structuring. It is often impractical to predict in advance what types of information need to be extracted and what corresponding tools should be utilized, limiting its adaptability to diverse and dynamic video analytics scenarios.

Alternatively, researchers have proposed obtaining relevant frames through multiple iterative retrieval processes~\cite{wang2024videoagent, wang2024videotree, ma2024drvideo, yang2024vca}. For instance, VideoAgent~\cite{wang2024videoagent} typically begins with a coarse-grained sampling of video segments to establish an initial high-level understanding. Based on this, the VLM is prompted to decide which finer-grained segments to retrieve and analyze in subsequent iterations.  However, these approaches face significant challenges when applied to video analytics scenarios involving extremely long videos. On one hand, the initial coarse-grained sampling may become insufficient as video length increases, potentially missing critical information. On the other hand, the iterative retrieval and analysis process becomes increasingly computationally expensive as video duration grows, making it impractical for large-scale video analytics tasks.

Recent studies have advanced RAG techniques~\cite{edge2024graphrag, lazygraphrag2024, guo2024lightrag, fan2025minirag} by incorporating knowledge graph construction to enhance the retrieval process. However, these works primarily focus on text-only RAG problems, and adapting such approaches to video analytics remains a significant challenge due to the complexity and multimodal nature of video data. In this paper, we propose \sysname, which, to the best of our knowledge, is the first system to enable VLM-powered video analytics by effectively addressing the aforementioned challenges.

%% file: tex/overview.tex
\section{\sysname System Overview}
\label{sec:overview}

The key idea of \sysname lies in leveraging a small VLM to efficiently structure video streams into discrete \emph{events}, then linking these events by extracting insights from each to construct a comprehensive index. Given a specific query, \sysname leverages this index to proactively retrieve relevant information from both the index and the associated raw frames. Ultimately, the retrieved data are utilized by the VLM to produce a coherent and contextually appropriate response.

To build \sysname as the L4 video analytics system, we established the following design principles: 1) The analytics should be scalable to any volume of video data, \ie exceeding hundreds of hours, while ensuring that the computational overhead remains independent of the video length; 2) The index construction must operate in near-real-time, allowing the system to support timely event analytics; 3) The system should accommodate not only fact-based retrieval queries but also query-focused summarization and multi-hop queries, supporting open-ended analytics.

To this end, as depicted in Fig.~\ref{fig:overview}, \sysname system is composed of two primary components: \emph{near-real-time index construction} and \emph{agentic retrieval and generation}. Within each component, we introduce a set of techniques designed to effectively realize the established design principles.

In the index construction phase, our objective is to design an effective index while ensuring construction efficiency. To achieve this, we introduce the \emph{event knowledge graph} (EKG) to structure video streams (\S\ref{subsec:event_knowledge_graph}). An EKG is a specialized form of a knowledge graph (KG) designed to represent and organize at the granularity of events and their interconnections. Recognizing that events unfold across varying temporal scales, we propose \emph{semantic chunking} (\S\ref{subsec: sematic_chunking}) to extract meaningful events from video streams. Specifically, video streams are segmented into small, uniform chunks (\eg 3-second intervals), and a small VLM, such as Qwen2.5-VL-7B, is periodically employed to generate detailed content descriptions for these chunks using carefully crafted prompts. Subsequently, neighboring chunks are merged into larger semantic chunks by identifying semantically equivalent textual descriptions with BERTScore~\cite{devlin2019bert}. For each semantic chunk, the small VLM extracts entities and their relationships. Identical entities across different events are linked to ensure consistency and coherence. Ultimately, \sysname facilitates the continuous construction of an EKG for a given video stream, regardless of its length, providing a comprehensive representation of semantic events, entities, and their interrelations in near-real-time on typical edge servers.

In the retrieval and generation process, we aim to leverage the constructed index for efficiently retrieving essential and minimal information, and to robustly generate the final answer based on the retrieved data. To achieve this, we first introduce the concept of \emph{tri-view retrieval}. Specifically, a given query undergoes simultaneous retrieval across three dimensions: events, entities, and visual embeddings. This approach ensures the acquisition of comprehensive and relevant information pertaining to the query. To further support complex queries in L4 video analytics systems \eg query-focused summarization and multi-hop queries, we propose an \emph{agentic searching} mechanism. In particular, by utilizing the LLM as an agent, \sysname proactively explores to retrieve additional information from events linked to those retrieved in earlier steps. \sysname explores multiple pathways to gather information and formulates a response to the query based on the collected data. Finally, we introduce the \emph{thoughts-consistency} strategy, which selects the most coherent and accurate final answer from multiple generated candidates.

%% file: tex/construction.tex
\section{Near-Real-Time Index Construction}
\label{sec:offline}

\subsection{Event Knowledge Graph}
\label{subsec:event_knowledge_graph}

An Event Knowledge Graph (EKG) is a structured representation of events and their interconnections, linking entities, timestamps, locations, and other contextual information to offer a holistic understanding of events and their dependencies. By employing an EKG, the content of a video can be organized into a sequence of events, associating groups of entities with specific events and capturing their intricate relationships.

Although existing works, such as GraphRAG~\cite{edge2024graphrag} and LightRAG~\cite{guo2024lightrag}, utilize knowledge graphs (KGs) to construct retrieval indices, we argue that EKGs are more suitable for video data. The rationale lies in the fundamental difference between the two: KGs focus on static entities (\eg people, locations, concepts) and their attribute-based relationships, whereas EKGs prioritize modeling dynamic events and their spatiotemporal evolution. 

\begin{figure}[t]
    \centering
    \includegraphics[width=0.85\linewidth]{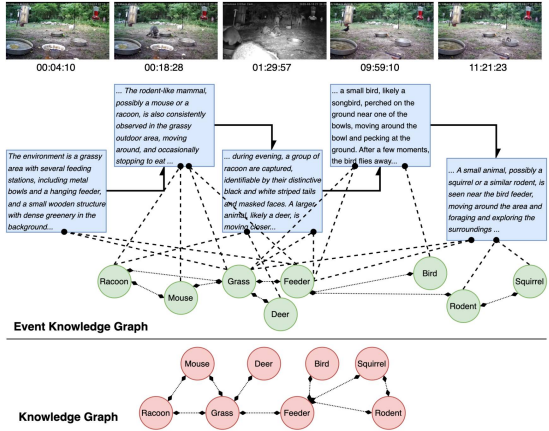}
    \caption{An example of a constructed event knowledge graph and a knowledge graph from wildlife monitoring scenarios for video analytics.}
    \label{fig:evg_kg}
\end{figure}

Fig.~\ref{fig:evg_kg} illustrates an example video alongside its corresponding KG and EKG. As depicted, the EKG effectively captures key events and their transitions, while representing entities with finer granularity within specific events. This enables EKG-based retrieval to support more sophisticated queries, such as event summaries, multi-hop temporal reasoning, and other complex analyses. In contrast, KGs, which only encapsulate entities across the entire video, lack the capability to fundamentally support such advanced queries.

Formally, we define our EKG $\mathcal{G}$ as follows: 
\begin{equation}
    \mathcal{G} = (\mathcal{E}, \mathcal{U}, \mathcal{R}),
\end{equation}
where $\mathcal{E} = \{e_i\}_{i=1}^{|\mathcal{E}|}$ represents the temporally ordered set of events, $\mathcal{U} = \{u_j\}_{j=1}^{|\mathcal{U}|}$ denotes the entities extracted from the video within each event, and $\mathcal{R} = \mathcal{R}_{ee} \cup \mathcal{R}_{uu} \cup \mathcal{R}_{ue}$ encompasses three types of relationships: 1) temporal event-event relations $\mathcal{R}_{ee}$, such as \emph{before} and \emph{after}, which encode temporal logic constraints; 2) semantic entity-entity relations $\mathcal{R}_{uu}$, akin to the relationships found in conventional KGs; and 3) participation relations $\mathcal{R}_{ue}$, which associate entities with their contextual roles within specific events.

\subsection{Semantic Chunking}
\label{subsec: sematic_chunking}

To construct the EKG as an index, it is essential to extract events and their corresponding descriptions from videos. Although current VLMs demonstrate remarkable capabilities in event detection and transcription, their application in video analytics scenarios remains challenging.

On one hand, large VLMs, such as Qwen2.5-VL-72B, can achieve high accuracy in event detection and transcription, but their substantial computational overhead makes it difficult to process video streams in near-real-time, particularly on resource-constrained edge servers~\cite{10.1145/3719664}. On the other hand, small VLMs, such as Qwen2.5-VL-7B, offer reduced latency but suffer from performance degradation as the length of the video increases. Furthermore, both large and small VLMs are limited by their constrained context windows. To handle long video content, a common approach is to partition the content uniformly, a process known as chunking. However, events in videos naturally occur at varying and diverse temporal scales. Inaccurate chunking can disrupt the coherence of individual events, thereby increasing the difficulty for VLMs to accurately detect and transcribe them.

To address this, we propose a semantic chunking approach. The core idea involves processing a video stream in the following steps: First, we perform uniform buffering, \eg dividing the video into fixed-length chunks of 3 seconds each. Next, a small VLM, \eg Qwen2.5-VL-7B, is employed to extract representative event descriptions from these chunks with proper prompts. Based on the generated event descriptions, we utilize a text embedding model, such as BERT~\cite{devlin2019bert}, to measure the similarity between neighboring events. Adjacent events with high similarity are then merged into a single event. Ultimately, this process enables \sysname to partition the entire video into semantically meaningful chunks while simultaneously extracting their corresponding descriptions.

\begin{figure}[t]
    \centering
    \includegraphics[width=0.8\linewidth]{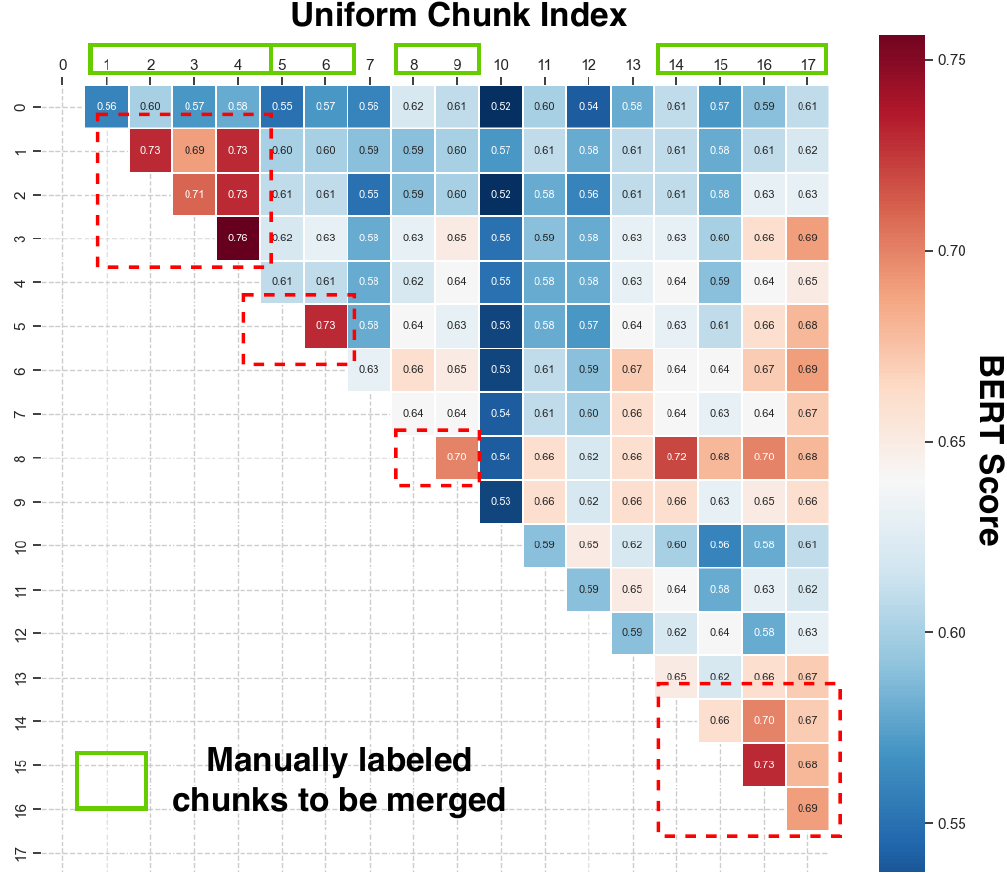}
    \caption{Merging uniform chunks into semantic chunks guided by the pairwise BERTScore distribution.}
    \label{fig:semantic_chunking}
\end{figure}

Particularly, an input video $V$ is initially divided into uniform chunks $c_i$, and a small VLM is employed to generate textual descriptions $d_i$ for each chunk $c_i$. Subsequently, the semantic similarity between any two uniform chunks is measured by computing the pairwise BERTScore~\cite{zhangbertscore} for $(d_i, d_j)$. Higher similarity scores suggest that the same event may occur across these chunks, making them candidates for semantic merging. Specifically, we adopt two criteria to determine whether certain uniform chunks can be merged into a single semantic chunk: 1) Within a semantic chunk, the similarity between any two uniform chunks must exceed a predefined threshold (\eg 0.65 in our implementation); 2) After merging, the similarity between the boundaries of adjacent semantic chunks must fall below a sufficiently low threshold. Fig.~\ref{fig:semantic_chunking} illustrates the semantic chunking process, where a video initially divided into 18 uniform chunks is successfully merged into 9 semantic chunks. Once merged, the small VLM is further utilized to summarize each semantic chunk.

It is important to highlight, although pairwise BERTScore computations are performed multiple times, \sysname efficiently schedules these computations in parallel, leveraging the hardware parallelism (\S\ref{sec:implementation}). Consequently, the semantic chunking process does not become a bottleneck in the near-real-time index construction phase, as detailed in \S\ref{sec:evaluation}.

\subsection{Entity Extraction and Linking}
\label{subsec:entity_extraction}

In addition to extracting event information from videos, \sysname also identifies entities and their relationships, as illustrated in Fig.~\ref{fig:overview}. Similar to the approach in~\cite{edge2024graphrag}, we utilize a small VLM to extract entities and their relationships from videos using carefully designed prompts for each event.

The identified entities, however, tend to be highly redundant across events within the EKG. Such redundancy not only increases storage requirements but also hampers retrieval efficiency. Thus, it is necessary to de-duplicate and link these entities. Existing works~\cite{edge2024graphrag, guo2024lightrag, fan2025minirag}, which primarily focus on text-only RAG problems, typically rely on exact string matching strategies for entity de-duplication. However, in the context of video analytics and EKG, entities are independently extracted from each event by the VLM, leading to potential inconsistencies in entity descriptions for the same concept across different events, \eg, "raccoon" and "procyon lotor".

To address this, \sysname employs a text embedding model \eg JinaCLIP~\cite{koukounas2024jinaclip}, to encode all extracted entities into vector representations. Using embedding similarities as a metric, we apply a standard K-means clustering algorithm to group entities. This approach ensures that semantically similar entities are de-duplicated and linked by forming unified clusters. 
\revised{To represent each cluster, we compute the centroid of the embedding vectors of all entities within that cluster, which serves as the representative feature of the merged entity.}

Ultimately, the constructed EKG is stored in a database comprising five tables: events, entities, event-to-event relationships, entity-to-entity relationships, and entity-to-event relationships. Additionally, the raw video frames are vectorized using JinaCLIP~\cite{koukounas2024jinaclip} and linked to their corresponding events, enabling comprehensive retrieval in the following phase.

%% file: tex/generation.tex
\section{Agentic Retrieval and Generation}
\label{sec:online}

In the agentic retrieval and generation stage, our primary objectives are to effectively retrieve relevant information using the constructed EKG and to generate robust, contextually accurate responses based on the retrieved data. To achieve this, we introduce an agentic searching mechanism that explores multiple retrieval pathways within the index to gather the necessary information. Additionally, we propose a thought consistency strategy, enabling \sysname to generate and evaluate multiple responses from different retrieval pathways, ultimately selecting the most appropriate one.

\subsection{Tri-View Retrieval}
\label{subsec:tri_view}

\begin{figure}
    \centering
    \includegraphics[width=0.9\linewidth]{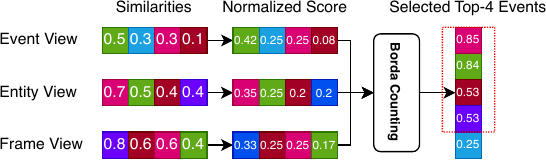}
    \caption{An illustration of tri-view retrieval and borda counting on the retrieved events.}
    \label{fig:borda_counting}
\end{figure}

To comprehensively retrieve relevant information from the index for a given query, \sysname employs a tri-view retrieval process: the first view targets events, enabling retrieval at the event level to provide information for event summary-related queries. Specifically, the query is encoded using the text encoder JinaCLIP~\cite{koukounas2024jinaclip} and matched against the events table in the constructed EKG. The second view focuses on entities, offering insights into basic facts or item-specific queries. For this, we leverage the entity centroids extracted and aggregated as detailed in \S\ref{subsec:entity_extraction} to facilitate retrieval. The third view utilizes vision embeddings of raw video frames as complementary information. The retrieved entities and raw frames are subsequently linked to their associated events through the constructed EKG.

It is important to note that the retrieved events should be ranked. Ranking is not only crucial for filtering noise from the retrieved results but also essential for enabling agentic searching, as detailed in \S\ref{subsec:agentc_searching}. A straightforward ranking method, such as similarity-based ranking, cannot be directly applied to \sysname due to the retrieved events originating from three distinct views. To integrate these results, we propose to use a weighted Borda counting approach.

Fig.~\ref{fig:borda_counting} illustrates the process of using Borda counting to integrate and rank retrieved events from the three views in \sysname. Specifically, we select the top $K$ events from each view and rank them based on their calculated similarities within that view. Subsequently, the similarities of these $K$ events are normalized to compute their Borda scores:

\begin{equation}
    s_{m}(e_j) = \frac{\text{sim}_m(e_j)}{\sum_{e_k \in \mathcal{E}_m} \text{sim}_m(e_k)},
\end{equation}
where $\mathcal{E}_m$ represents the set of events retrieved from view $m$. The final Borda score for each event $e_j$ is then obtained by summing its scores across all views:

\begin{equation}
    s(e_j) = \sum_{m} s_m(e_j),
\end{equation}

Finally, the aggregated Borda scores $s(e_j)$ are used to rank all retrieved events.

\begin{figure}
    \centering
    \includegraphics[width=0.9\linewidth]{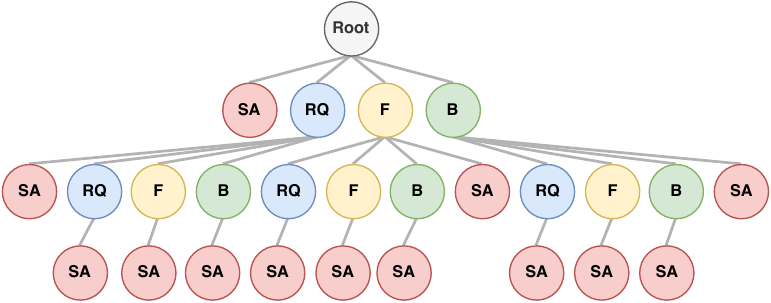}
    \caption{An example of agentic tree search with four actions and a depth of three, yielding 13 distinct pathways for information gathering and response generation.}
    \label{fig:tree_search}
\end{figure}

\subsection{Agentic Searching on Graph}
\label{subsec:agentc_searching}

The retrieved events mentioned above can be directly utilized to generate the final answer. However, to support complex queries, such as query-focused summaries and multi-hop queries, \sysname searches for additional relevant information by leveraging the relationships between events and entities within the constructed EKG. To enable efficient exploration, we propose the agentic searching on graph approach.

\revised{
The proposed approach is inspired from human strategies for information retrieval and reasoning within video content. Typically, individuals initiate this process by locating salient segments guided by retrieval keywords (\ie query), subsequently enriching their understanding by examining the temporal context in adjacent clips (\ie forward and backward). For a more thorough comprehension, they may iteratively refine their search using alternative keywords as necessary (\ie re-query). Similarly, in our agentic searching process, we define the agentic action space as follows:
}

\textbf{Forward (F)}: this action extends the current retrieval by including temporally subsequent events on the EKG for all events in the event list. It reflects the natural tendency of humans to seek forward narrative progression when trying to understand what happens next or how a situation evolves over time.
    
\textbf{Backward (B)}: complementing the forward action, this action retrieves temporally preceding events, enabling a backward exploration of the narrative to uncover prior context or causal factors.
    
\textbf{Re-query (RQ)}: this action generates a new query represented by a list of keywords via an LLM and retrieves complementary events as outlined in \S\ref{subsec:tri_view}. It reflects the human tendency to gather information from multiple perspectives to achieve a more comprehensive understanding.

\textbf{Summary and Answer (SA)}: this action utilizes the descriptions of the retrieved events from the EKG and generates the response to the specific query by employing an LLM.

\revised{
Utilizing these predefined agentic actions, we structure the agentic search process within a tree search framework. The search is initiated by performing an initial retrieval based on the original query, yielding a set of relevant events that constitute the root node of the search tree, as outlined in \S\ref{subsec:tri_view}. For each node, a single-step rollout is conducted in which four predefined actions, namely, \textit{forward}, \textit{backward}, \textit{re-query}, and \textit{summary and answer}, are executed on every events on the node. Upon reaching the {SA} action, the corresponding search trajectory is terminated by generating an answer. The rollout proceeds iteratively until the predefined maximum tree depth is reached. This tree search mechanism systematically explores multiple pathways to extract information from the EKG. As depicted in Fig.~\ref{fig:tree_search}, a tree of depth 3 yields 13 distinct information-gathering paths along with their respective answers. All generated answers are subsequently evaluated, with the optimal answer selected via the thought consistency method elaborated in \S\ref{subsec:thoughts_consistency}.
}

A practical issue in the tree search process is the exponential growth in the number of retrieved events as the tree depth increases. This not only introduces computational overhead but also results in the accumulation of noisy or irrelevant information. To mitigate this, we use a length constraint on the maintained event list during the search process, \ie 16 in our implementation. When the number of retrieved events exceeds this limit, we employ a drop strategy to discard less relevant events based on their rankings described in \S\ref{subsec:tri_view}.

\subsection{Consistency Enhanced Generation}
\label{subsec:thoughts_consistency}

During the agentic tree search, multiple candidate answers are generated at {SA} nodes across different pathways. To determine the final answer, it is necessary to either select or synthesize from these candidates. A straightforward approach would be majority voting. However, due to the diversity of retrieval paths, only a small subset of these nodes is likely to access the essential information with minimal noise, producing high-quality answers. To this end, we introduce the thoughts-consistency mechanism to identify and select the most reliable final answer.

At each SA node, instead of generating the answer a single time, we repeatedly generate answers multiple times using a Chain-of-Thought (CoT) prompting scheme. Following the principle of self-consistency~\cite{wang2023self}, correct answers are more likely to emerge consistently across multiple valid reasoning trajectories during repeated generations. Specifically, we evaluate the consistency not only across the generated answers but also within their associated CoT traces. Nodes demonstrating strong internal coherence, where the reasoning process aligns logically with the conclusion, are assigned higher scores.

To formalize this process, we propose a scoring framework that integrates both \emph{answer agreement} and \emph{thought consistency}. At each SA node, we perform $n$ rounds of sampling using a temperature setting between $0.5$ and $0.7$, resulting in a set of $n$ candidate outputs denoted as $\{(a_i, r_i)\}_{i=1}^n$, where $a_i$ is the answer and $r_i$ is the associated reasoning trace.

Let $\mathcal{A} = \{a^{(1)}, a^{(2)}, \dots, a^{(T)}\}$ be the set of unique answers among the $n$ samples, where $T$ is the number of distinct answers. The answer agreement score $S_a^{(t)}$ for a candidate answer $a^{(t)}$ is defined as the proportion of times it appears in the samples:
\begin{equation}
S_a^{(t)} = \frac{|\{i \mid a_i = a^{(t)}\}|}{n}
\end{equation}

The thought consistency score $S_r^{(t)}$ for $a^{(t)}$ is computed as the average BERTScore between all pairs of reasoning traces associated with $a^{(t)}$:
\begin{equation}
S_r^{(t)} = \frac{2}{k(k-1)} \sum\limits_{1 \leq i < j \leq k} \text{BERTScore}(r_i, r_j),
\end{equation}
where $k$ is the number of times $a^{(t)}$ appears in the $n$ samples. The final score for each candidate answer combines these two components:
\begin{equation}
\label{eq:thoughts_consistency_weight}
S_{\text{final}}^{(t)} = \lambda S_a^{(t)} + (1 - \lambda) S_r^{(t)},
\end{equation}
where $\lambda \in [0, 1]$ is a weighting parameter controlling the trade-off between answer agreement and thought consistency. In our implementation, we set it to 0.3, the parameter tuning would be discussed in \S\ref{sec:thoughts_consistency}.

For each {SA} node, the candidate answer with the highest $S_{\text{final}}^{(t)}$ is selected as its definitive response. To enhance the reliability of this final answer, we propose an additional agentic action, \textbf{Check Frames and Answer (CA)}. This action retrieves the raw video frames associated with the events from the EKG and utilizes the VLM to generate a refined response to the specific query. By doing so, this action effectively supplements any missing information relevant to the query that may have been overlooked during the construction phase.

Specifically, after ranking all candidate answers from the {SA} nodes using the consistency-enhanced scoring mechanism, the top-2 nodes with differing answers are selected. The video frames corresponding to their retrieved events are extracted, and the VLM is prompted to generate a new response by directly attending to the visual evidence. Furthermore, the thought-consistency mechanism is applied to the {CA} nodes to bolster the reliability of the final generated answer.

%% file: tex/implementation.tex
\section{Implementation}
\label{sec:implementation}

In \sysname{} we use Qwen2.5-VL-7B for constructing EKGs, Qwen2.5-32B for SA, and Gemini-1.5-Pro for CA. \revised{We utilize AWQ~\cite{lin2024awq} and LMDeploy~\cite{2023lmdeploy} to accelerate the on-device inference. These particular LLMs are selected for \sysname due to our two-step principle: we first determine an appropriate model size that is feasible for edge deployment, and then, within that size range, choose the best-performing model available according to the public benchmarks~\cite{fu2024videomme, shen2024longvu}.} Additionally, \sysname{} adopts batch inference for several key stages—including description generation, description merging, entity extraction, and tree search—to improve efficiency and maximize GPU utilization. For text and vision embedding, we utilize JinaCLIP\cite{koukounas2024jinaclip}.
And we employ BERTScore with the deberta-xlarge-mnli\cite{he2020deberta} checkpoint. The storage of EKG and vector representations is based on the implementation of \cite{guo2024lightrag}, upon which we make further modifications to suit the specific requirements of \sysname{}. 

During the EKGs construction stage, we carefully design prompts to guide the extraction of structured information from video. For general-purpose video understanding, we employ a unified prompt that avoids introducing bias or prior assumptions:

\textit{"...Your task is to extract and provide a detailed description of the video segment, focusing on all key visible details..."}. 

For scenario-specific videos, we design the prompts to emphasize scenario-relevant information. \revised{For example, in the case of wildlife Monitoring scenario, key information may includes the timestamps of the recording, animal activities (\eg presence, species, number, specific behaviors, etc.), and environmental changes. Particularly the prompts used in \sysname are listed in \S\ref{app:prompts}}

\revised{The hyperparameters in \sysname are determined through a parameter tuning procedure conducted on the selected datasets. Specifically, for the semantic chunking methodology described in \S\ref{subsec: sematic_chunking}, we set the BERTScore threshold for merging and summarizing descriptions at 0.65. Empirically, this threshold yielded partitions that exhibited greater alignment with human annotations across various scenarios. Additionally, the maximum tree search depth is set to 3, more details are provided in \S\ref{subsub:depth}.}

%% file: tex/evaluation.tex
\section{Evaluation}
\label{sec:evaluation}

\subsection{Evaluation Settings}

\begin{figure*}[t!]
    \centering
    \begin{subfigure}[t]{0.32\textwidth}
        \centering
        \includegraphics[width=\linewidth]{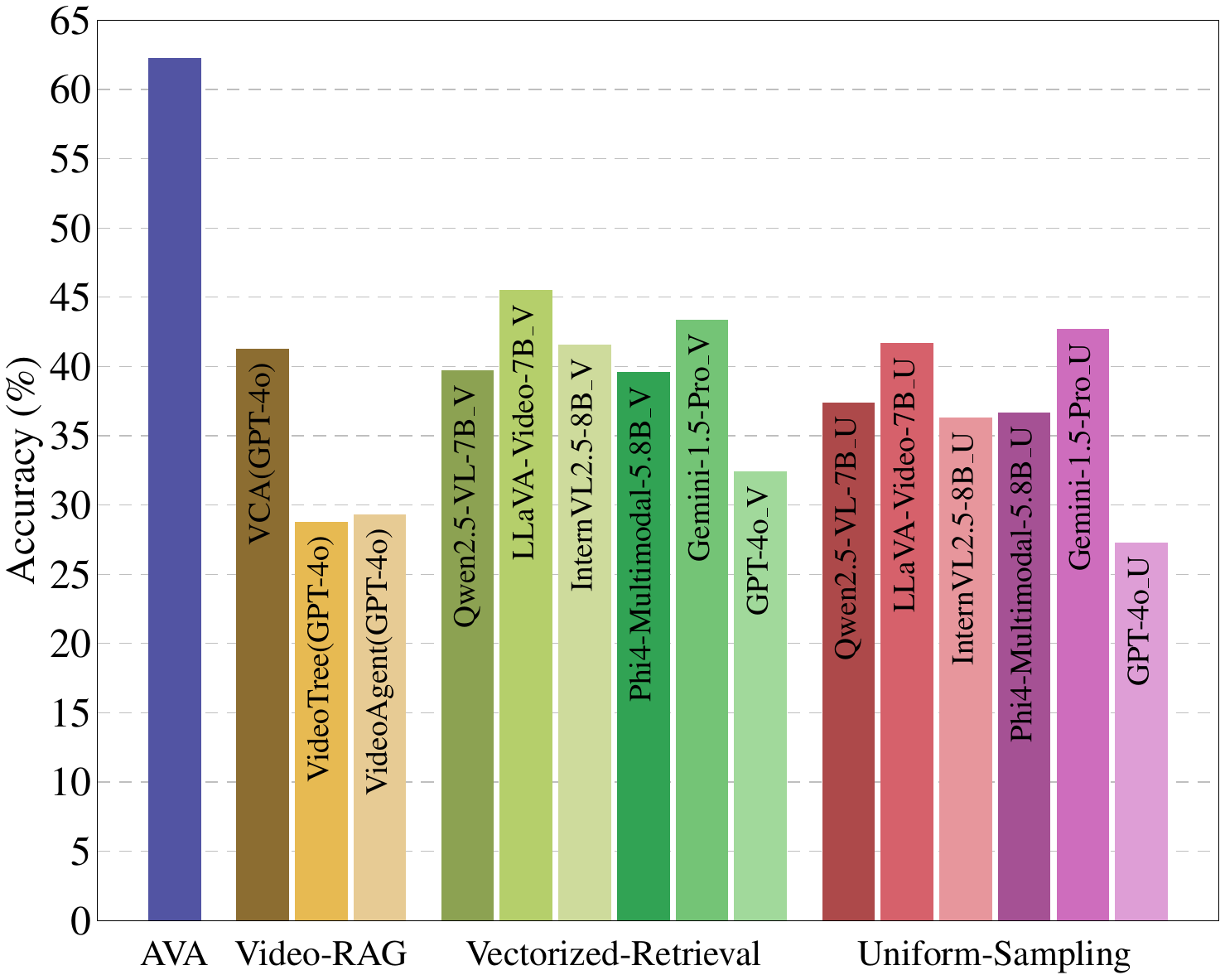}
        \caption{LVBench}
        \label{subfig:overall_performance_lvbench}
    \end{subfigure}
    \hfill
    \begin{subfigure}[t]{0.32\textwidth}
        \centering
        \includegraphics[width=\linewidth]{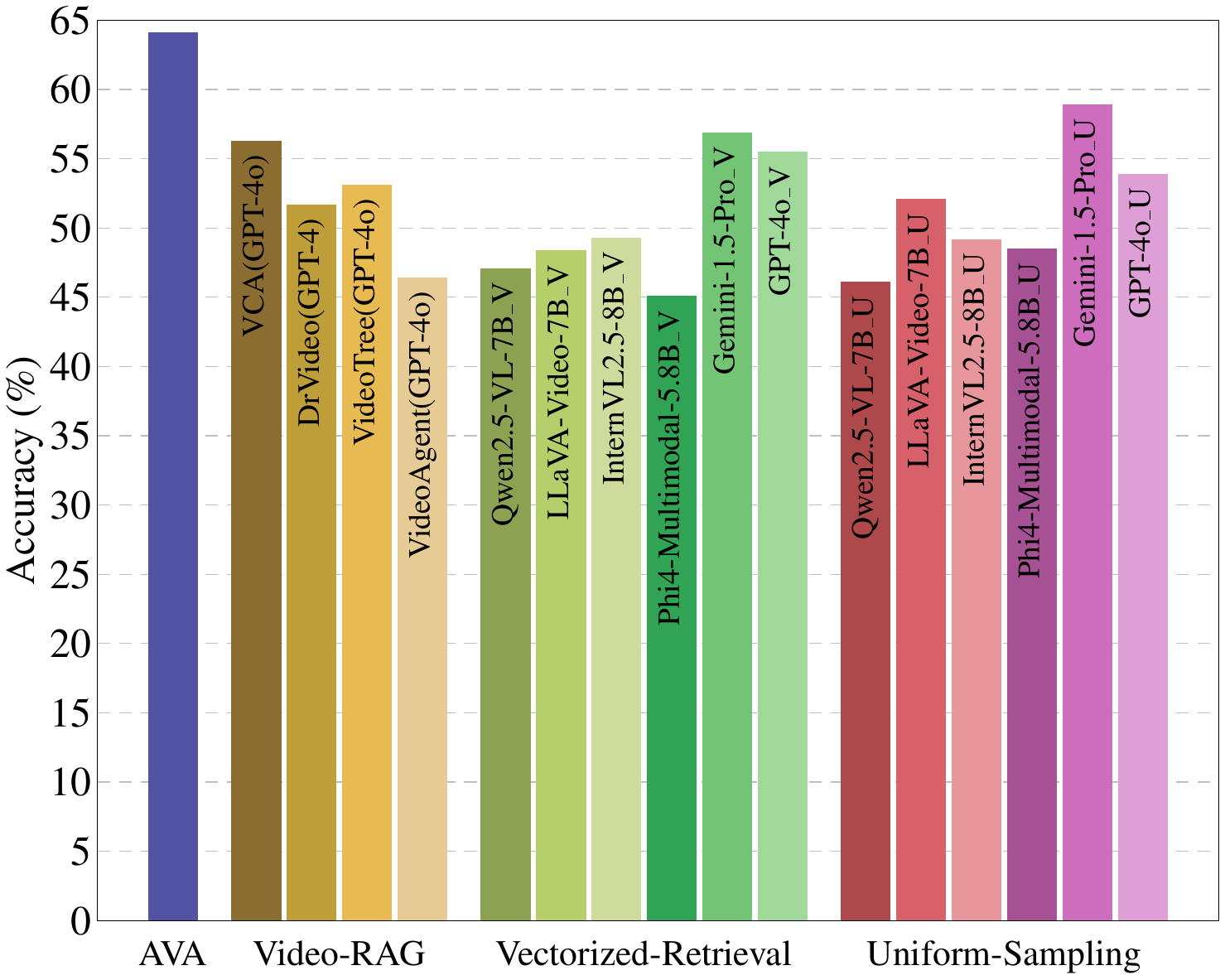}
        \caption{VideoMME-Long}
        \label{subfig:overall_performance_videomme}
    \end{subfigure}
    \hfill
    \begin{subfigure}[t]{0.32\textwidth}
        \centering
        \includegraphics[width=\linewidth]{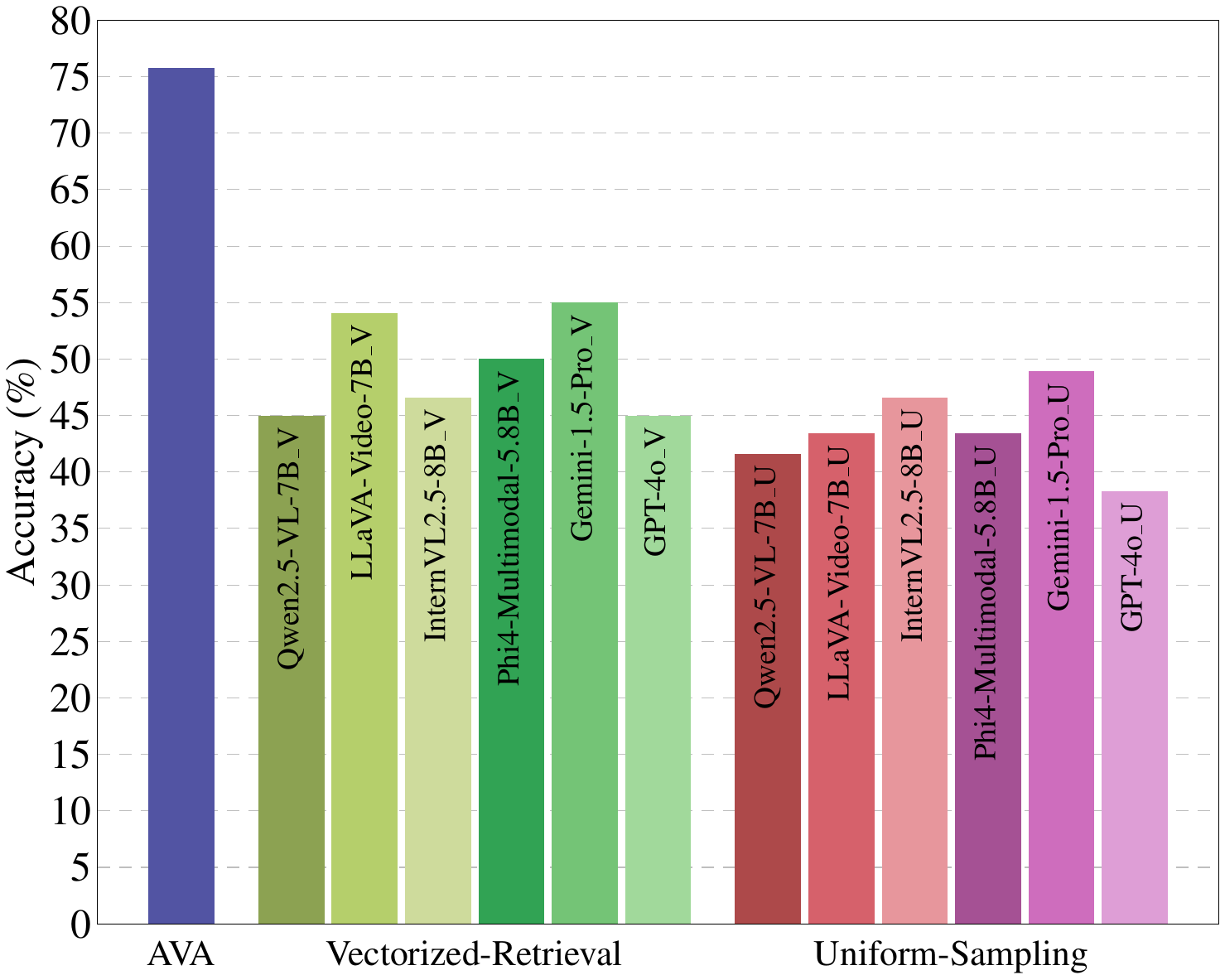}
        \caption{\sysname{}-100}
        \label{subfig:overall_performance_ava100}
    \end{subfigure}
    \vspace{-1mm}
    \caption{The achieved accuracy of \sysname{} and various baselines on the LVBench, VideoMME-Long, and \sysname{}-100 benchmarks.}
    \label{fig:overall_acc}
    \vspace{-2mm}
\end{figure*}

\subsubsection{Benchmarks}
\sysname{} is evaluated on two widely used public long-video benchmarks and one ultra-long video benchmarks proposed by us, covering a broad range of video scenarios and problem types.

\textbf{LVBench}\cite{wang2024lvbench} stands out among publicly available benchmarks for its exceptionally long average video duration, approximately 4100 seconds per video. It comprises 103 videos with a total of 1549 questions, covering six distinct video domains and addressing six task types including temporal grounding, summarization, and reasoning.

\textbf{VideoMME-Long}\cite{fu2024videomme} is a subset of the VideoMME benchmark, focusing on videos exceeding 20 minutes in duration, with an average length of 2400 seconds. Comprising a total of 300 videos and 900 questions, the benchmark covers a wide range of video themes, spanning 6 primary visual domains with 30 subfields to ensure broad scenario generalizability, and includes 12 distinct task types.

\textbf{\sysname{}-100} is proposed by us, which is an ultra-long video benchmark specially designed to evaluate video analysis capabilities
\sysname{}-100 consists of 8 videos, each exceeding 10 hours in length, and includes a total of 120 manually annotated questions. The benchmark
covers four typical video analytics scenarios: human daily activities, city walking, wildlife monitoring, and traffic monitoring, each scenario contains two videos. The human daily activity scenario features egocentric videos selected and stitched from the Ego4D\cite{grauman2022ego4d}. City walking and wildlife monitoring videos are curated from publicly available recordings on YouTube, capturing urban exploration and animal monitoring respectively. Traffic monitor videos are composed from clips in the Bellevue Traffic Video Dataset~\cite{bhardwaj2022ekya}. All questions are carefully designed by human annotators, who also provide reference answers as the ground truth. In addition, GPT-4o is utilized to generate plausible distractor options. The accuracy is evaluated by analyzing \sysname's responses to multiple-choice questions included in the benchmarks. \revised{More details regarding \sysname-100 are provided in \S\ref{app:benchmark}.}

\subsection{Baselines}
We conduct a comprehensive comparison between \sysname{} and a wide range of baseline models, encompassing both mainstream VLMs and specialized Video-RAG methods. The VLM baselines include GPT-4o~\cite{achiam2023gpt}, Gemini-1.5-Pro~\cite{team2023gemini}, Phi-4-Multimodal~\cite{abdin2024phi4}, Qwen2.5-VL-7B~\cite{bai2025qwen2_5}, InternVL2.5-8B~\cite{chen2024internvl}, and LLaVA-Video-7B~\cite{zhang2024llavavideo}. Each of these models is evaluated with two typical strategies: uniform sampling and vectorized retrieval, where a CLIP-based retriever selects the top-K relevant frames based on the user query. In addition to VLMs, we benchmark \sysname{} against SOTA Video-RAG frameworks, including VideoTree\cite{wang2024videotree}, VideoAgent\cite{wang2024videoagent}, DrVideo\cite{ma2024drvideo}, and VCA\cite{yang2024vca}. Among these, VideoTree, VideoAgent, and VCA are built upon GPT-4o, while DrVideo leverages GPT-4.

\begin{figure}[!h]
    \centering
    \includegraphics[width=0.85\linewidth]{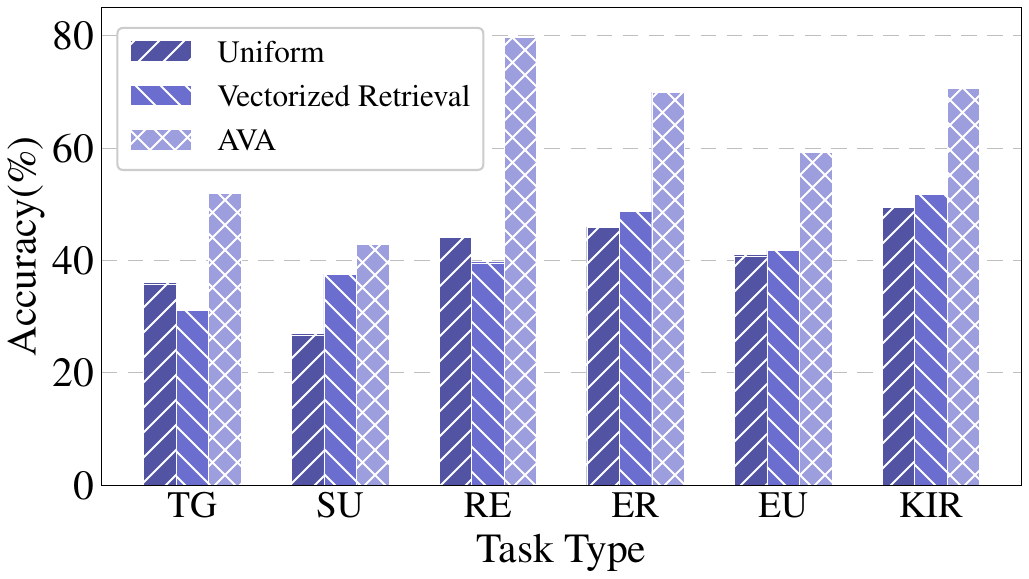}
    \caption{The accuracy achieved by \sysname{} and the baselines across typical query categories on LVBench: Temporal Grounding (TG), Summarization (SU), Reasoning (RE), Entity Recognition (ER), Event Understanding (EU), and Key Information Retrieval (KIR).\vspace{-0.25cm}}
    \label{fig:task_specific}
\end{figure}

\subsection{Overall Evaluation}

\subsubsection{Overall Performance}

Fig.~\ref{fig:overall_acc} illustrates the overall accuracy achieved by \sysname{} compared to various baselines on the LVBench, VideoMME-Long, and \sysname{}-100 benchmarks. Across all three benchmarks, \sysname{} consistently outperforms the baselines. Specifically, on LVBench, \sysname{} delivers a remarkable 16.9\% improvement, while on VideoMME-Long, it advances the SOTA by approximately 5.2\%. On the \sysname{}-100 benchmark, \sysname{} achieves an accuracy of 75.8\%, significantly surpassing all competing methods.

In detail, compared to video-RAG methods, \sysname{} achieves improvements of 21\% and 7.8\% on LVBench and VideoMME-Long, respectively. When compared to vectorized retrieval-based methods, \sysname{} demonstrates gains of 16.9\% on LVBench and 20.8\% on \sysname{}-100. Furthermore, against uniform sampling baselines, \sysname{} improves performance by approximately 19.6\% and 26.9\% on LVBench and \sysname{}-100, respectively.

Notably, on \sysname{}-100, when evaluated with extremely long videos, \sysname{} maintains robust performance, whereas the baselines degrade significantly. This highlights the effectiveness of \sysname{} in handling L4 video analytics tasks.

\subsubsection{Performance on Different Query Categories}
We also evaluate the accuracy achieved by \sysname{} across typical query categories on LVBench. As illustrated in Fig.~\ref{fig:task_specific}, our approach achieves improvements of 16\%, 5.3\%, 35.6\%, 21.2\%, 17.5\%, and 18.9\% across six key task types: Temporal Grounding, Summarization, Reasoning, Entity Recognition, Event Understanding, and Key Information Retrieval, respectively, compared to the uniform sampling and vectorized retrieval baselines powered by Gemini-1.5-Pro. Notably, \sysname{} demonstrates particularly strong performance on reasoning tasks, which require identifying causal relationships between events and linking preceding and succeeding events within the video. This highlights \sysname{}'s ability to effectively locate and extract critical information from long videos, thereby enabling advanced L4 video analytics systems.
\begin{figure}[!t]
    \centering
    \includegraphics[width=0.85\linewidth]{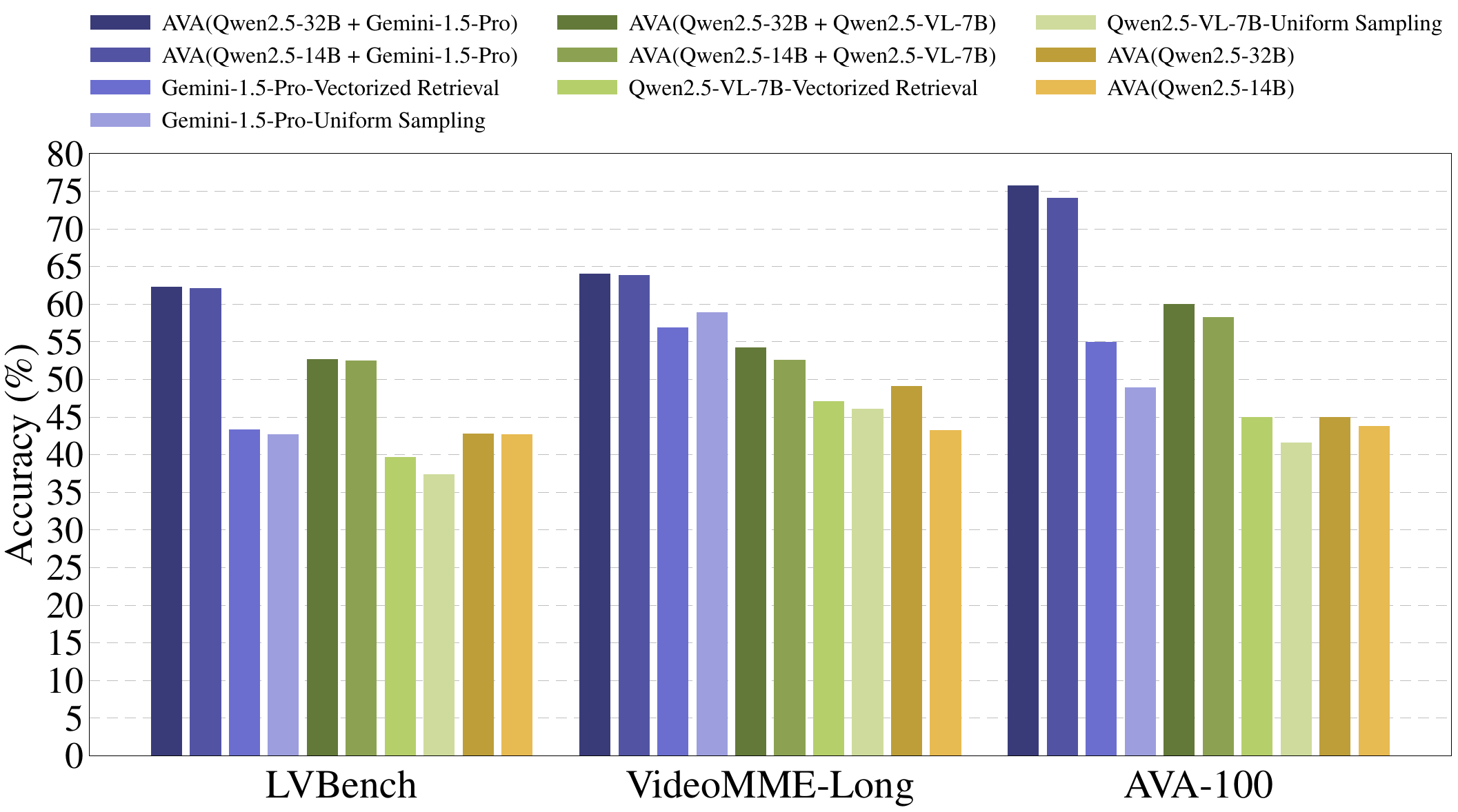}
    \caption{The accuracy achieved by \sysname{} and baselines across three benchmarks when utilizing different LLMs and VLMs.}
    \label{fig:ava_configs}
\end{figure}

\subsubsection{Performance under Different Configurations}
Fig.~\ref{fig:ava_configs} shows the performance of \sysname{} using different models configurations for SA and CA. For SA, two models were used: Qwen2.5 14B and 32B. For CA, two models were used: Qwen2.5-VL-7B and Gemini-1.5-Pro. 
The results show that across the three benchmarks, \sysname{} using Gemini-1.5-Pro for CA achieved improvements of 18.9\%, 5.2\%, and 20.8\% respectively compared to the best baseline result using the same model, while using Qwen2.5-VL-7B yielded improvements of 13\%, 7.2\%, and 15\% respectively, fully demonstrating the effectiveness of our method. Notably, even when only using Qwen2.5-32B and Qwen2.5-7B based on the textual content from EKG without accessing raw frames, \sysname{} can surpass the performance of Qwen2.5-VL-7B on the three benchmarks and also outperform most models shown in Figs.~\ref{subfig:overall_performance_lvbench}, \ref{subfig:overall_performance_videomme}, and \ref{subfig:overall_performance_ava100}.

\subsubsection{Performance on Different Video Lengths}

\begin{figure}[t]
    \centering
    \resizebox{0.8\linewidth}{!}{\includegraphics{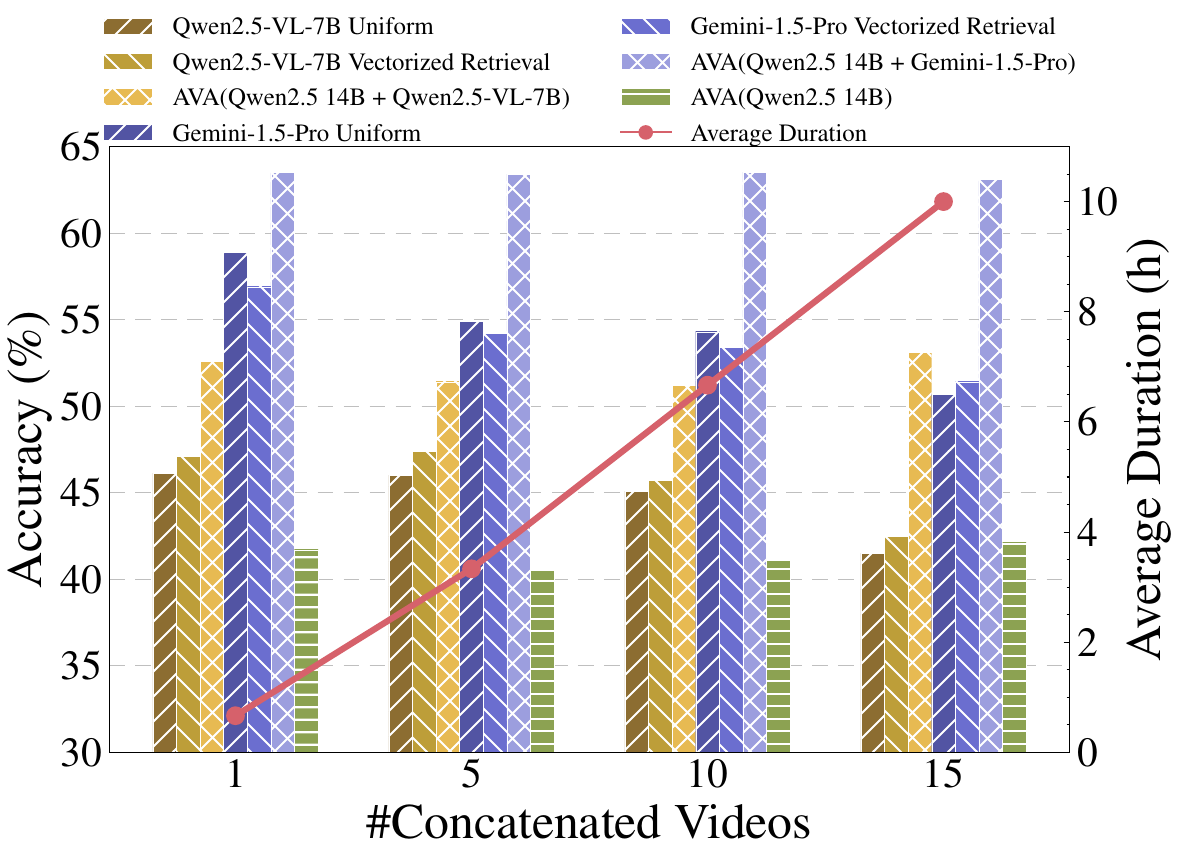}}
    \caption{The accuracy achieved by \sysname{} and the baselines across varying video lengths via concatenating videos from LVBench.}
    \label{fig:concat}
\end{figure}

To evaluate the robustness of \sysname{} with respect to video length, we conducted experiments on videos of varying durations. Specifically, sequences of 3.3, 6.6, and 10 hours were created by concatenating videos from the VideoMME-Long benchmark. Performance was measured using identical questions across these varying video lengths. As illustrated in Fig.~\ref{fig:concat}, both Qwen2.5-VL-7B and Gemini-1.5-Pro baselines exhibit significant performance degradation as video length increases. When extended to 10 hours, their performance declines by 4.6\% and 8.2\%, respectively, under the uniform sampling method, compared to the original VideoMME-Long benchmark. For the vectorized retrieval setting, the performance drops are 4.6\% and 5.5\%, respectively. These results highlight the limitations of these methods in scaling effectively with increasing video length. In contrast, \sysname{} consistently maintains stable performance across all video lengths, underscoring its robustness and scalability in handling video data of any duration.

\subsubsection{System Overhead}

\begin{figure}[t]
    \centering
    \resizebox{0.85\linewidth}{!}{\includegraphics{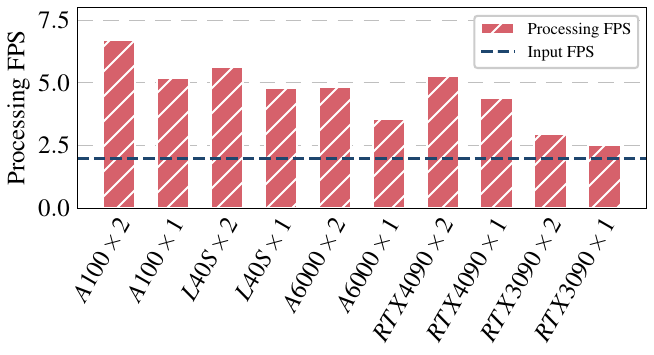}}
    \caption{Total index construction overhead evaluated on various types of typical edge server hardware.}
    \label{fig:construction overhead}
\end{figure}

\sysname{} is designed to enable the near real-time construction of EKGs. As shown in Fig.~\ref{fig:construction overhead}, we measured the average processing speed (in FPS) of \sysname{} while constructing EKGs from LVBench videos across various hardware platforms, with the input video stream fixed at 2 FPS. On 2 $\times$ A100 GPUs, \sysname{} achieved an impressive processing speed of 6.7 FPS, significantly exceeding the input stream rate. On a single RTX 4090, a typical edge server hardware, \sysname{} maintained a processing speed of 4.4 FPS, still well above the input frame rate. Even on a single RTX 3090, \sysname{} performed effectively, achieving 2.5 FPS. This performance demonstrates its capability to support efficient, near real-time EKG construction for L4 video analytics. The overhead during the retrieval and generation phases will be discussed further in \S\ref{subsec:ablation}.

\revised{
To evaluate system overhead during the generation phase, we conducted a detailed latency analysis of three stages within the generation pipeline, as summarized in Table~\ref{tab:generation_overhead}. The experiment is performed on one A100 GPU with LMDeploy~\cite{2023lmdeploy}.~\footnote{\revised{We configure $cache\_max\_entry\_count$ to 0.3 in LMDeploy, restricting the KV cache memory consumption to no more than 30\% of the total GPU memory.}} The retrieval stage utilizing JinaCLIP demonstrates notable efficiency, requiring only 0.44 seconds per query and consuming less than 1 GB of GPU memory on average. The agentic search stage presents the predominant source of latency: 101.5 seconds with Qwen2.5-14B, rising to 174.2 seconds with the larger Qwen2.5-32B, thereby underscoring the inference time of LLMs as the principal runtime overhead. During the consistency-enhanced generation stage, Qwen2.5-VL-7B incurs a latency of 45.8 seconds and approximately 31 GB of memory usage, whereas the API-based Gemini-1.5-Pro achieves substantially lower latency at 14.2 seconds. The results demonstrate that the agentic searching stage represents the principal performance bottleneck, underscoring the importance of strategic model selection to balance accuracy and computational efficiency. Furthermore, the results also highlight the imperative of optimizing the agentic searching process, which we identify as a key direction for future research.
}

\begin{table}[!t]
\centering
\resizebox{0.9\linewidth}{!}{
\begin{tabular}{cccc}
\toprule[1.5pt]
\textbf{Stage} & \textbf{Model} & \textbf{Latency (s)} & \textbf{GPU Memory (GB)} \\
\midrule
Tri-View Retrieval & JinaCLIP & 0.44 & 0.8 \\
\midrule
\multirow{2}{*}{Agentic Searching} & Qwen2.5-14B & 101.5 & 30 \\
 & Qwen2.5-32B & 174.2 & 40 \\
\midrule
\multirow{2}{*}{Consistency Enhanced Gen.} & Qwen2.5-VL-7B & 45.8 & 31 \\
 & Gemini-1.5-Pro & 14.2 & - \\ 
\bottomrule[1.5pt]
\end{tabular}
}
\caption{\revised{The breakdown of latency and GPU memory consumption across the three stages of the generation phase in \sysname. All measurements are performed on a single A100 GPU, with LLMs deployed via LMDeploy~\cite{2023lmdeploy} utilizing AWQ~\cite{lin2024awq}.}}
\label{tab:generation_overhead}
\end{table}

\subsection{Ablation Evaluation}
\label{subsec:ablation}
We randomly sampled 20 videos and 305 corresponding questions from LVBench for our ablation study. All ablation experiments were conducted on 2 $\times$ A100 GPUs.

\begin{table}[!t]
\centering
\label{tab:context_length}
\begin{tabular}{lcc} 
\toprule[1.5pt]
\textbf{Method} & \textbf{Acc.} & \textbf{Overhead(h)}  \\ 
\midrule
MiniRAG         &     28.1          &          3.49                   \\
LightRAG        &   30.6            &   3.52                          \\
\sysname{}         &      \textbf{39.7}         &       \textbf{0.31}                    \\
\bottomrule[1.5pt]
\end{tabular}
\caption{The achieved accuracy and construction overhead evaluated when using EKG and KG as index in \sysname{} and baseline models on the subset of LVBench. The total video duration is around 1.2 hours.}
\label{tab:ekg_kg}
\end{table}

\subsubsection{Different Index Construction Methods}
We compare \sysname{}’s EKG construction method with two representative knowledge graph-based 
construction methods: LightRAG~\cite{guo2024lightrag} and MiniRAG~\cite{fan2025minirag}. Since both of them only support text-only construction, we use the full set of descriptions obtained through the semantic chunking (\S\ref{subsec: sematic_chunking}) as their input textual corpus. 
We use Qwen2.5 7B to construct EKG and KG for \sysname and baselines, respectively. In the retrieval and generation phase, we use the same LLM, Qwen2.5 14B and the same settings, \eg maximum number of tokens of retrieved events or entities.

As shown in Table~\ref{tab:ekg_kg}, \sysname{} demonstrates a significant performance advantage over the baselines. Specifically, it achieves 11.6\% higher accuracy than MiniRAG and 9.1\% higher accuracy than LightRAG. Crucially, this improved performance comes with substantially less construction overhead, requiring only 0.31 hours compared to 3.49 and 3.52 hours for the baselines. The rationale is that baselines construct KG based on massive uniform chunks, while \sysname utilizes the semantic chunks. This substantial gap in both effectiveness and efficiency highlights that \sysname{}'s EKG construction method not only yields higher-quality knowledge representations but also drastically reduces the time needed to build the graph.

\subsubsection{Different Tree Search Depths}

We also evaluate the achieved accuracy and overhead applying different settings in the retrieval phase, \ie tree search depth. The effect of tree depth lies in a crucial trade-off: while shallower depths may struggle to retrieve comprehensive information, increasing the depth allows access to richer information from deeper nodes. However, this comes with a significant increase in tree search overhead, and the information from deeper levels can introduce more noise, potentially negatively impacting the final generation quality. Table~\ref{tab:tree depth} presents the results comparing different tree search depths on performance and tree search overhead. As shown, performance generally increases with increasing tree depth up to a certain point. Specifically, for all three \sysname{} configurations evaluated, the highest accuracy is achieved at a tree search depth of 3. Accuracy decreases when the depth is further increased to 4, suggesting that excessive depth leads to the retrieval of detrimental noise or irrelevant information, outweighing the benefit of additional context. Conversely, the tree search overhead increases sharply with depth. Expanding the search from depth 1 (6.7s) to depth 2 (27.3s) incurs a moderate increase. 
Comparing the accuracy improvements and the overhead increase, a tree search depth of 3 offers the optimal balance. 

\label{subsub:depth}
\begin{table}[t!]
\centering
\resizebox{\linewidth}{!}{
\begin{tabular}{lcccc} 
\toprule[1.5pt]
\multicolumn{1}{c}{\multirow{2}{*}{\textbf{Method}}} & \multicolumn{4}{c}{\textbf{Tree Search Depth}}  \\
\multicolumn{1}{c}{}                                 & 1    & 2    & 3             & 4                 \\
\midrule
\sysname{}(Qwen2.5 14B)                                   & 34.1 & 36.1 & \textbf{40.9} & 39.5              \\
\sysname{}(Qwen2.5 14B + Qwen2.5VL 7B)                    & 49.3 & 52.1 & \textbf{53.8} & 50.2              \\
\sysname{}(Qwen2.5 14B + Gemini-1.5-Pro)                  & 54.2 & 58.4 & \textbf{61.5} & 52.7              \\
\midrule[1pt]
\multicolumn{1}{c}{\textbf{Tree Search Overhead(s)}}               & 6.7 & 27.3 & 90.1 & 370.3 \\
\bottomrule[1.5pt]
\end{tabular}
}
\caption{The achieved accuracy and overhead when applying different tree search depths in the agentic search of \sysname evaluated on the subset of LVBench.}
\vspace{-1em}
\label{tab:tree depth}
\end{table}

\subsubsection{Different Consistency Evaluation Settings}
For consistency-enhanced generation, \sysname{} incorporates two key parameters: $\lambda$, which governs the balance between the contributions of thought consistency and answer consistency, and the number of generations for self-consistency evaluation. Fig.~\ref{subfig:self_consistency_weight} illustrates the impact of varying $\lambda$ values on the accuracy achieved by \sysname{}. Notably, the optimal performance is observed when $\lambda$ is set to $0.3$, highlighting the importance of jointly considering both intermediate thought consistency and final answer consistency to ensure robust results. As depicted in Fig.~\ref{subfig:self_consistency_times}, the accuracy of \sysname{} gradually improves as the number of self-consistency iterations increases. However, this improvement comes at the expense of significantly higher computational overhead. For example, increasing the self-consistency iterations from 8 to 16 yields only a 0.9\% accuracy gain, while nearly doubling the computational cost. This demonstrates a clear trade-off between marginal accuracy improvements and resource efficiency. Balancing this trade-off, we adopt 8 self-consistency iterations in the implementation of \sysname{}, ensuring a practical balance between performance and computational overhead.

%% file: tex/discussion.tex
\section{Limitations and Future Work}

There are also limitations in the current design of \sysname, which we explore for future work. Specifically: 1) The existing agentic retrieval and generation mechanism relies on a fixed tree-search strategy based on the Monte Carlo approach. While effective, this method is computationally expensive. The trajectories collected during the search process could be leveraged as training data to develop a model capable of dynamically selecting optimal search actions and depths based on the query and context. 2) Although the integrated VLM demonstrates robust general video understanding and reasoning capabilities, it may encounter challenges in certain specialized visual tasks, such as precise object counting. Incorporating lightweight, task-specific vision models as tools within the system could improve accuracy for such queries. Our future work will focus on enabling the VLM, functioning as an autonomous agent, to intelligently invoke these specialized tools, thereby addressing its limitations in handling specific tasks.

%% file: tex/conclusion.tex
\label{sec:thoughts_consistency}
\begin{figure}[!t]
    \centering
    \begin{subfigure}[t]{0.48\linewidth}
        \centering
        \includegraphics[width=\linewidth]{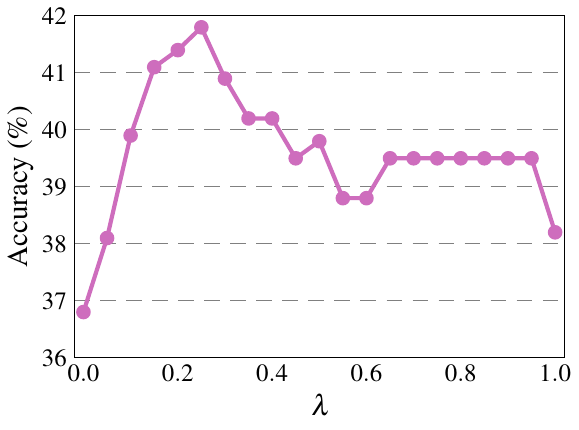}
        \caption{Balance between thoughts and answer consistency}
        \label{subfig:self_consistency_weight}
    \end{subfigure}
    \hfill
    \begin{subfigure}[t]{0.48\linewidth}
        \centering
        \includegraphics[width=\linewidth]{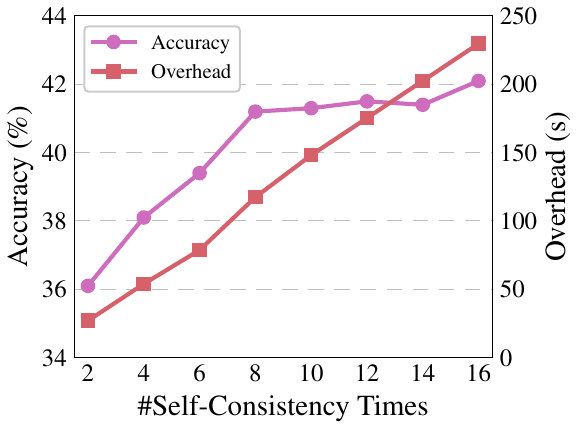}
        \caption{Trade-offs using different self-consistency times}
        \label{subfig:self_consistency_times}
    \end{subfigure}
    \vspace{-1mm}
    \caption{Performance of \sysname{} under varying consistency evaluation settings on a subset of LVBench.}
    \vspace{-2mm}
    \label{fig:overall_performance}
\end{figure}
\section{Conclusion}

This paper presents \sysname, an advanced L4 video analytics system powered by VLMs. \sysname enables comprehensive understanding and open-ended query analysis of large-scale, long-duration video data, overcoming the constraints of existing video analytics systems that are predominantly tailored to specific, pre-defined tasks. The system introduces novel designs, including near-real-time Event Knowledge Graph index construction and an agentic retrieval and generation mechanism, facilitating efficient organization and analysis of extended video content to address complex queries. We demonstrate \sysname's superior performance on public video understanding benchmarks, as well as on our newly proposed benchmark, \sysname-100, specifically designed to evaluate video analytics tasks.

%% file: tex/acknowledgement.tex
\section*{Acknowledgements}
This work is partially supported by NSFC under grant No. 92467301, Key Research and Development Program of Zhejiang Province (Grant No: 2025C01012), and the ZJUCSE-Enflame cloud and edge intelligence joint laboratory.

%% file: tex/appendix.tex
\appendix

\revised{
\section{\sysname-100 Benchmark}
\label{app:benchmark}
\subsection{Benchmark overview}
\label{app:benchmark_overview}
\begin{figure}[h]
    \centering
    \resizebox{0.95\linewidth}{!}{\includegraphics{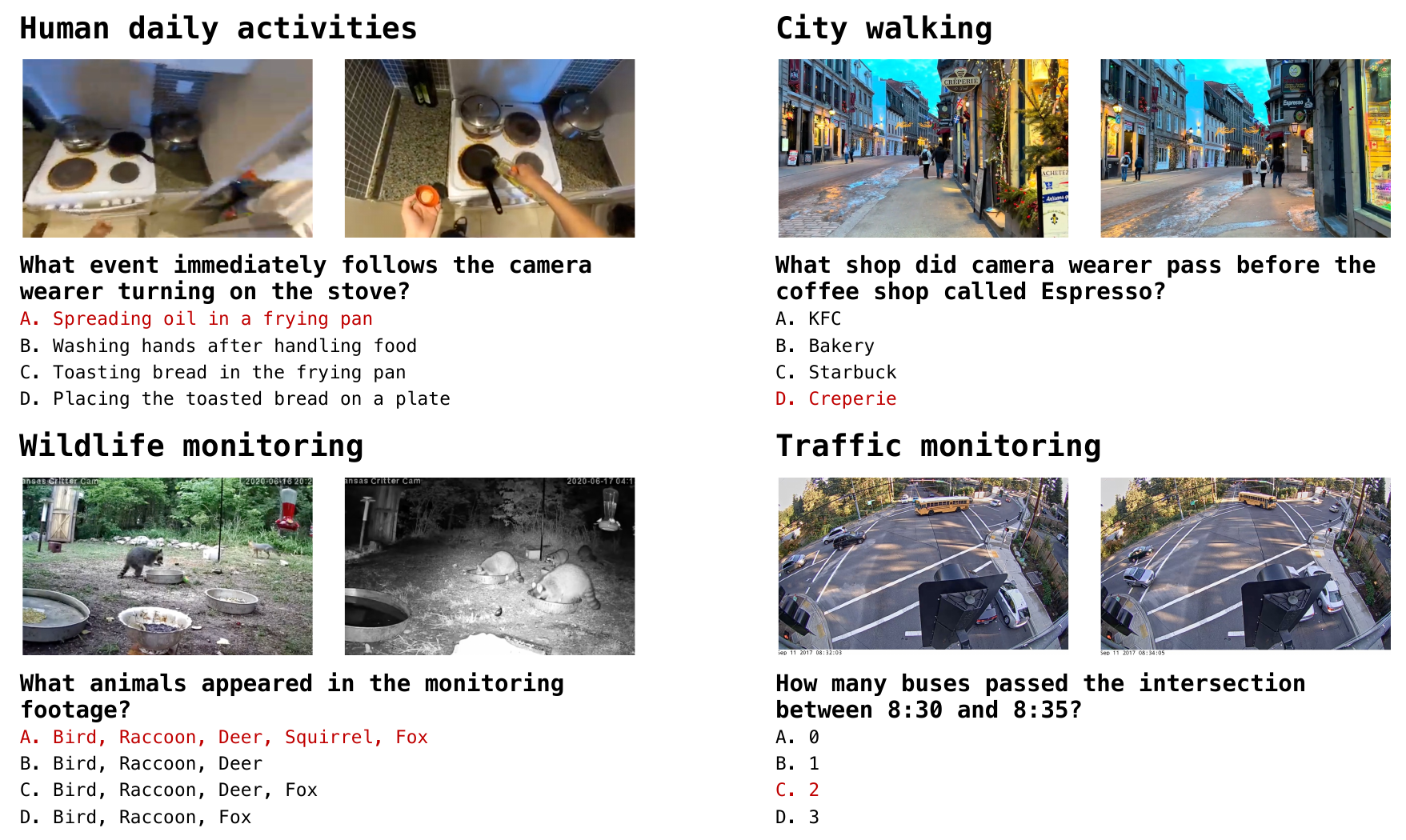}}
    \caption{\revised{Scenarios and QA examples in \sysname-100.}}
    \label{fig:ava-100-overview}
\end{figure}

\sysname-100 is established as a benchmark to advance research in L4 video analytics systems, with a particular focus on ultra-long-duration video understanding and complex reasoning tasks. In contrast to existing datasets that primarily focus on short video clips or domain-specific activities, \sysname-100 centers on real-world, continuous video streams often spanning several hours. This design enables the evaluation of video analytics systems under realistic and challenging scenarios, including long-horizon temporal reasoning, sparse event detection, and cross-segment summarization, etc.

The benchmark currently comprises eight ultra-long videos encompassing diverse scenarios including egocentric recordings of daily activities, city tours, as well as third-person fixed-camera capturing traffic monitoring and wildlife monitoring. Each video is annotated with multiple question–answer (QA) pairs: human annotators generate ground-truth QAs, while GPT-4o is employed to produce multiple-choice options, subsequently verified through manual review. In total, \sysname-100 contains approximately 100 hours of video content and 120 annotated QA pairs, offering a valuable resource for the evaluation of both video-language models and specialized video analytics systems.

Fig.~\ref{fig:ava-100-overview} illustrates the representative scenarios and QA examples featured in \sysname-100. Table~\ref{tab:ava100} presents detailed statistics of the \sysname-100 dataset, encompassing video duration and the number of QA pairs. The benchmark is publicly available at \url{https://huggingface.co/datasets/iesc/Ava-100}.
}

\revised{
\subsection{Selected Scenarios and Data Sources}
\subsubsection{Human Daily Activities}

The egocentric daily-life scenario videos in \sysname-100 are primarily sourced from the Ego4D dataset~\cite{grauman2022ego4d}. Ego4D is one of the largest existing egocentric video datasets, containing more than 3,600 hours of video collected across multiple countries, focusing on a wide range of daily activities such as household routines, cooking, shopping, and social interactions. It was originally designed to support research in egocentric perception, including tasks like action recognition, episodic memory, forecasting, and hand–object interaction understanding.  

\begin{table}[h]
\centering
\resizebox{\linewidth}{!}{
\begin{tabular}{lccc}
\toprule[1.5pt]
\textbf{Video ID} & \textbf{Duration (hours)} & \textbf{\#QA Pairs} & \textbf{Views} \\
\midrule
ego-1 & 12.7 & 22 & First-person (moving) \\
ego-2 & 11.7 & 19 & First-person (moving) \\
citytour-1 & 12.0 & 19 & First-person (moving) \\
citytour-2 & 10.5 & 20 & First-person (moving) \\
\midrule
traffic-1 & 14.9 & 12 & Third-person (fixed) \\
traffic-2 & 13.9 & 13 & Third-person (fixed) \\
wildlife-1 & 12.0 & 8  & Third-person (fixed) \\
wildlife-2 & 11.5 & 7  & Third-person (fixed) \\
\midrule
\textbf{Total} & \textbf{99.2} & \textbf{120} & -- \\
\bottomrule[1.5pt]
\end{tabular}
}
\caption{\revised{Statistics of the \sysname-100 dataset, including video duration, number of QA pairs, and perspective attributes for each video. The first four entries correspond to egocentric videos capturing moving perspectives, whereas the last four represent fixed third-person recordings.}}
\label{tab:ava100}
\end{table}

Ego4D does not typically provide single, continuous clips exceeding 10 hours in duration. To construct long-form egocentric videos for \sysname-100, we employed the concatenation strategy. Specifically, we prioritized the preservation of event diversity over maintaining the identity of the camera wearer. Accordingly, we selected and concatenated sub-clips from distinct Ego4D videos to generate extended sequences surpassing 10 hours. Throughout this process, we guarantee that the concatenated sub-clips encompassed a wide spectrum of human activities and contexts, thereby enhancing the richness of the video content while mitigating potential annotation ambiguities.
}

\revised{
\subsubsection{City Walking}

The first-person city tour videos in \sysname-100 are sourced from the YouTube channel \textit{4K World Wandering}\cite{4KWorldWandering}, which features high-resolution walking tour videos recorded in various locations around the globe.  These videos capture natural human navigation experiences in public spaces, making them a valuable complement to the daily-life egocentric content. The videos typically span up to four hours in length. To construct long-form sequences for \sysname-100, we adopted the same concatenation strategy as described above. Specifically, we aimed to compose extended videos by combining multiple sub-clips, while preserving both geographic and contextual coherence.  

In particular, we curated sub-videos predominantly from the same region or country to construct each long-form video. Meanwhile we intentionally enhanced diversity by integrating segments filmed under various weather conditions (\eg sunny, snowy, rainy) and at different times of day (\eg daytime, night) into the concatenated sequences. This design enriches the video contents, rendering the city walking videos in \sysname-100 more representative of real-world navigation scenarios across diverse geographic and temporal conditions.

The city walking videos prioritize large-scale spatial cognition, long-term trajectory tracking, and environmental variability. Consequently, they present added challenges for QA annotation, as the questions frequently require reasoning about landmarks, navigation routes, or temporal dynamics within the environment.
}

\revised{
\subsubsection{Traffic Monitoring}

The traffic monitoring videos in \sysname-100 are sourced from the Bellevue Traffic dataset~\cite{bhardwaj2022ekya}, which comprises continuous, fixed-camera recordings of real-world traffic scenes. This dataset is particularly well-suited for the study of long-term temporal dynamics, vehicle-pedestrian interactions, and congestion analysis. 

For \sysname-100, we selected two intersections: Bellevue\_150th\_Newport (spanning from 2017-09-11 03:08:29 to 2017-09-11 16:08:32) and Bellevue\_Bellevue\_NE8th (spanning from 2017-09-11 03:08:29 to 2017-09-11 15:08:32). These extensive, continuous recordings encompass a broad spectrum of traffic events across different periods of the day, including both peak hours and quieter intervals, thereby providing a rich and diverse dataset.

To construct QA pairs, we target the fine-grained traffic events, such as vehicle movements, pedestrian crossings, and congestion incidents. For example, whether a red car passed through the intersection between 4:30 and 4:40 and in which direction it traveled; the presence of pedestrians or cyclists at specific times. Such temporally anchored, detail-oriented QAs highlight the unique value of traffic monitoring videos for evaluating long-duration reasoning.

}

\revised{
\subsubsection{Wildlife Monitoring}

The wildlife monitoring videos in \sysname-100 are sourced from publicly available YouTube live-streaming channels, which continuously capture activities in natural habitats through fixed outdoor cameras. Particularly, we select four sources. The first source is the Arkansas Critter Cam Channel\cite{ArkansasCritterCam}, from which we used the video titled "Wildlife Live – African Waterhole 24/7", originally published on June 17, 2020. This stream features footage from a wildlife camera located in Arkansas, USA.  The second source is the "Nature Live Cams" channel\cite{NatureLive}, which specializes in streaming diverse natural environments worldwide, including forests, rivers, and savannahs. We selected the video "African Safari Live 24/7 – Watering Hole and Savannah" and extracted two segments from its April 12, 2025 livestream: 05:03–11:03 and 17:49–23:49. These segments were concatenated to form one ultra-long video. We try to ensure the coverage of animal activities across different times of the day, from early morning to late night. It is worth noting that wildlife scenes are inherently random and unpredictable, resulting in key events occurring infrequently and at indeterminate times within long-duration videos, posing additional challenges for L4 video analytics systems.
}

\revised{
\subsection{Prompts}
\label{app:prompts}

In \sysname-100, we employ a scenario-specific prompt design strategy, predicated on the view that prompts should be tailored to the deployment contexts of video analytics systems in order to optimize VLM performance by extracting salient information under varying conditions. For instance, prompts for first-person daily life videos are tailored to emphasize object interactions and activity sequences, whereas prompts for traffic monitoring footage prioritize vehicle trajectories, traffic flow, and pedestrian dynamics. Crucially, we regard prompt optimization as an integral component of system-level optimization within a fully VLM-driven video analytics framework. Thoughtfully crafted prompts can substantially mitigate redundant reasoning, enhance the efficiency of long-range temporal information extraction, and thereby elevate overall system performance. 

The prompts employed in our experiments are detailed as follows.

}

\balance

\clearpage
\begin{figure*}[t]
\centering
\begin{minipage}{0.9\textwidth}
\begin{lstlisting}[language=, caption={Description generation prompt for Human Daily Activities Scenario}, basicstyle=\footnotesize]
You are an expert in video understanding and description generation. 
You are given a first-person perspective video, and your task is to generate a continuous, smooth, and grounded description of the video content.

Focus particularly on:
- The actions and events performed by the camera wearer (the person holding or wearing the camera).
- The surrounding environment, including objects, people, and notable visual changes.
- The physical characteristics and spatial relationships of objects in the environment (e.g., size, color, relative positions, proximity to the camera wearer).
- Interactions between the camera wearer and the environment, including object manipulations and movements through space.

Avoid describing each frame individually, such as "frame1...". Instead, provide a coherent and logically structured narrative that flows smoothly over time.

Important constraints:
- Do not include assumptions, inferences, or fabricated details that are not visually evident.
- Do not speculate about the identity, emotions, or intentions of the camera wearer unless explicitly shown.
- When referring to the person holding or wearing the camera, always use the term "camera wearer".

Return your response as a single, continuous, and fluent paragraph that comprehensively describes the video content, including fine-grained visual details, and limit the length to 400 words.
\end{lstlisting}
\end{minipage}
\end{figure*}

\begin{figure*}[t]
\centering
\begin{minipage}{0.9\textwidth}
\begin{lstlisting}[language=, caption={Description generation prompt for City Walking Scenario}, basicstyle=\footnotesize]
You are an expert in video understanding and detailed scene description. 
You are given a first-person perspective video of a person walking through a city environment and your task is to generate a continuous, smooth, and grounded description of the video content.

Focus particularly on:
- The locations and landmarks the camera wearer passes by, such as buildings, shops, streets, and public spaces.
- The appearance and functions of these places (e.g., a small bakery with a red awning, a tall glass office building, a busy intersection).
- Events or notable occurrences observed during the walk, such as street performances, traffic changes, or people interacting in public.

Avoid describing each frame individually. Instead, provide a logically structured narrative that flows naturally over time.

Important constraints:
- Do not include assumptions, inferences, or fabricated details that are not visually evident.
- Do not speculate about the identity, emotions, or intentions of the camera wearer or other people unless explicitly shown.
- When referring to the person holding or wearing the camera, always use the term "camera wearer".

Return your response as a single, continuous, and fluent paragraph that comprehensively describes the video content, with attention to fine-grained urban and visual details, and limit the length to 400 words.
\end{lstlisting}
\end{minipage}
\end{figure*}

\begin{figure*}[t]
\centering
\begin{minipage}{0.9\textwidth}
\begin{lstlisting}[language=, caption={Description generation prompt for Traffic Monitoring Scenario}, basicstyle=\footnotesize]
You are a video analysis expert specializing in traffic observation and detailed event description. You are analyzing a road or intersection monitoring video recorded by a fixed-position camera.
Your task is to generate an **accurate, grounded, and coherent** description of the video segment.
Please focus on the following aspects:
- **Observed Traffic Elements:** Identify all traffic-related elements present in the video. Provide an integrated description covering the entire segment:
- **Vehicle Types:** Identify types as accurately as possible (e.g., car, truck, bus, motorcycle, bicycle, van). If unclear, describe the vehicle's physical characteristics (e.g., "a large box truck," "a small passenger vehicle," "a two-wheeled vehicle").
- **Quantity:** Indicate the number of each identified vehicle type, as well as the number of pedestrians.
- **Characteristics:** If relevant to the scene, note distinguishing physical features (e.g., color, size, presence of trailers, specific structural features). Describe pedestrians based on their interaction with traffic (e.g., walking along the sidewalk, crossing the street).
- **Actions / Events:** Describe observed dynamic behaviors and interactions (e.g., driving in a specific lane, stopping, turning, entering/exiting the frame, changing lanes, overtaking, pedestrians waiting or crossing), including any **traffic anomalies** (e.g., sudden braking, erratic maneuvers, red-light violations, collisions, traffic violations, illegal parking that obstructs traffic).
- **Timestamps:** Identify the timestamp shown on the monitoring footage.
**Output Format:**
After watching the full video segment, write a structured summary paragraph in the following format:
[Timestamp]: [Summary of vehicle types, quantities, characteristics, actions, pedestrian activity, and traffic anomalies].
**Important Constraints:**
- Do not include assumptions or details not clearly visible in the footage.
- Do not speculate about the intentions or emotions of drivers or pedestrians; only describe observable actions and postures.
- Use neutral descriptions such as "the footage shows" or "the camera captures"; avoid subjective phrasing like "we see" or "the viewer can observe."
- If vehicle type identification is uncertain, state this clearly.
- The final output should be a concise, fact-based summary of the traffic activity and scene context. The length should be appropriate to the events observed with a target length of less than 400 words.
\end{lstlisting}
\end{minipage}
\end{figure*}

\begin{figure*}[t]
\centering
\begin{minipage}{0.9\textwidth}
\begin{lstlisting}[language=, caption={Description generation prompt for Wildlife monitoring Scenario}, basicstyle=\footnotesize]
You are an expert in video analysis, specializing in wildlife observation and detailed environmental description.  
You are analyzing fixed-camera monitoring footage capturing a scene in a wild or natural environment and your task is to generate a precise, grounded, and chronologically ordered description of the entire video segment.

Focus on the following aspects:
- **Observed Animals:** Identify any animals present in the footage. For the entire segment, provide a consolidated description of:
- **Species:** Identify species as accurately as possible. If uncertain, describe physical characteristics (e.g., "a large brown bear", "a small rodent-like mammal", "a flock of unidentified birds").
- **Number:** Indicate the number of individuals observed.
- **Appearance:** Note distinctive physical features (e.g., size, color, antlers, markings).
- **Behavior:** Describe observed behaviors (e.g., foraging, running, resting, entering/exiting the frame, interacting).
- **Timestamps:** Identify the timestamp displayed in the monitoring footage.
- **Environment:** Briefly describe the static environment visible to the camera (e.g., forest clearing, rocky terrain, vegetation), and note any significant changes during the observation period (e.g., lighting shifts, weather changes).

**Output Format:**  
After reviewing the full segment, summarize your findings in a single structured paragraph using the following format:

[Timestamp]: [Environment description][Summary of animal and their activities]

**Important Constraints:**
- Do not include assumptions or invented details that are not visually evident in the footage.
- Do not speculate on the intentions or emotions of the animals; describe only observable actions and postures.
- Refer to observations using neutral terms such as "the footage shows" or "the camera captures"; avoid subjective phrasing like "we see" or "the viewer can observe".
- If species identification is uncertain, explicitly state this.
- The final output should be a concise, fact-based summary of the wildlife activity and environmental context of the segment, with a target length of less than 400 words.
\end{lstlisting}
\end{minipage}
\end{figure*}